\documentclass[a4paper,fleqn]{cas-sc}

\usepackage[authoryear,longnamesfirst]{natbib}

\usepackage{multirow}
\usepackage{amsmath}
\usepackage{verbatim}
\usepackage{color}
\usepackage{array}
\usepackage{graphicx}
\usepackage{subfigure}
\usepackage{amsmath,bm}
\usepackage{caption}
\usepackage{picinpar}

\begin{document}
\let\WriteBookmarks\relax
\def\floatpagepagefraction{1}
\def\textpagefraction{.001}
\shorttitle{Graph Partitioning and GNN based Graph Similarity Computation}
\shortauthors{Yueyang Wang, et~al.}

\title [mode = title]{Graph Partitioning and Graph Neural Network based Hierarchical Graph Matching for Graph Similarity Computation}

\author[1,2]{Haoyan Xu}
\cormark[1]

\author[1,2]{Ziheng Duan}
\cormark[1]

\author[2]{Jie Feng}
\cormark[1]

\author[2]{Runjian Chen}

\author[3]{Qianru Zhang}

\author[4]{Zhongbin Xu}

\author[1]{Yueyang Wang}[orcid=https://orcid.org/0000-0003-3210-0930]
\ead{yueyangw@cqu.edu.cn}
\cormark[2]

\address[1]{School of Big Data and Software Engineering, Chongqing University, Chongqing 401331, China}
\address[2]{College of Control Science and Engineering, Zhejiang University, Zhejiang 310027, China}
\address[3]{College of Foreign Languages, Harbin Institute of Technology, Heilongjiang 150001, China}
\address[4]{College of Energy Engineering, Zhejiang University, Zhejiang 310027, China}

\cortext[cor1]{Equal contribution with order determined by flipping a coin.}
\cortext[cor2]{Correspondence to: Yueyang Wang.}

\begin{abstract}
Graph similarity computation aims to predict a similarity score between one pair of graphs to facilitate downstream applications, such as finding the most similar chemical compounds similar to a query compound or Fewshot 3D Action Recognition. Recently, some graph similarity computation models based on neural networks have been proposed, which are either based on graph-level interaction or node-level comparison. However, when the number of nodes in the graph increases, it will inevitably bring about reduced representation ability or high computation cost.

Motivated by this observation, we propose a graph partitioning and graph neural network-based model, called PSimGNN, to effectively resolve this issue. Specifically, each of the input graphs is partitioned into a set of subgraphs to extract the local structural features directly. Next, a novel graph neural network with an attention mechanism is designed to map each subgraph into an embedding vector. Some of these subgraph pairs are automatically selected for node-level comparison to supplement the subgraph-level embedding with fine-grained information. Finally, coarse-grained interaction information among subgraphs and fine-grained comparison information among nodes in different subgraphs are integrated to predict the final similarity score.
Experimental results on graph datasets with different graph sizes demonstrate that PSimGNN outperforms state-of-the-art methods in graph similarity computation tasks using approximate Graph Edit Distance (GED) as the graph similarity metric.
\end{abstract}


\maketitle

\section{Introduction}

Graph similarity computation, which predicts a similarity score between one pair of graphs, has been widely used in various fields, such as recommendation system \cite{wu2015friend, hu2020graph}, computer vision \cite{horaud1989stereo, pelillo1999matching} and so on.
However, most of the standard distance measures evaluating how similar two graphs are, like Graph Edit Distance (GED) \cite{bunke1983distance}, and Maximum Common Subgraph (MCS) \cite{bunke1998graph}, still suffer from large search spaces or excessive memory requirements. They are weak to compute exact graph distance for graphs with more than 16 nodes \cite{blumenthal2018exact}.
Traditional graph similarity computation methods such as \textit{A*} \cite{riesen2013novel}, \textit{Hungarian} \cite{kuhn1955hungarian,riesen2009approximate}, \textit{VJ} \cite{fankhauser2011speeding,jonker1987shortest}, and \textit{Beam}\cite{neuhaus2006fast}, try to use pruning strategy or find approximate values instead of exact similarity to alleviate the problem.
Nevertheless, by performing directly from the graphs' edges and node characteristics, these exact and approximate algorithms still have a high time-complexity for computing the GED or MCS between two graphs and are hard to be generalized to large graphs in real applications.

With the rapid development of deep learning technology, graph neural networks that automatically extract the graph's structural characteristics provide a new solution for similarity computation and matching of graph structures. Recently, researchers proposed some representative graph deep learning models for graph similarity computation. During the training stage, these models fit the similarity ground truth (label) in a supervised learning way and learn a mapping between a pair of graph inputs and the similarity score. Hence they are more time-efficient compared with traditional graph similarity computation methods during testing or actual applications \cite{bai2019simgnn}.

\begin{figure}
\centering
\includegraphics[width=0.9\linewidth]{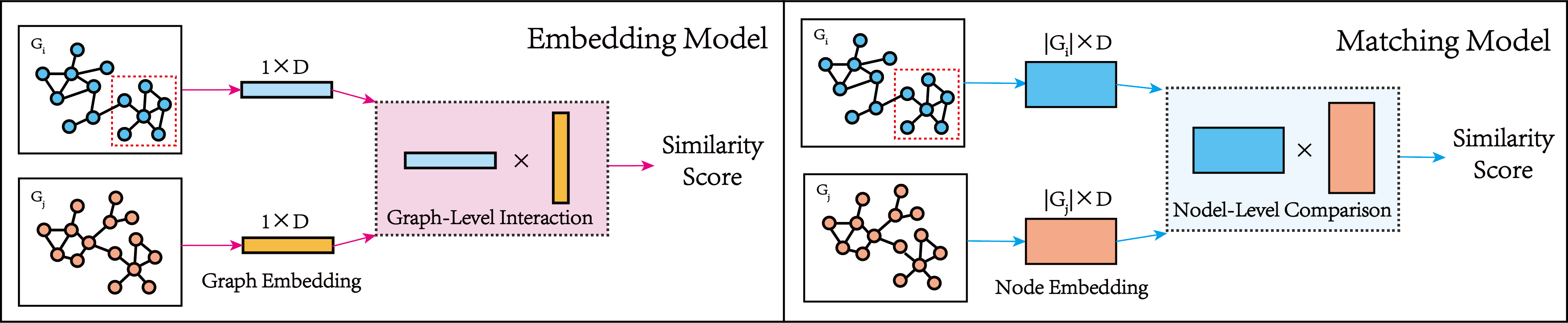}
\caption{Two deep learning frameworks for graph similarity computation: embedding-model and matching-model. $|G_i|$ and $|G_j|$ represent the number of nodes of $G_i$ and $G_j$, respectively, and $D$ represents the embedding dimension. The red dashed box in $G_i$ represents the subgraph of the $G_i$.}
\label{fig:1}
\vspace{-5mm}
\end{figure}

In general, graph deep learning models can be categorized into two classes, namely embedding model and matching model (shown in Fig \ref{fig:1}). \emph{Embedding-models} (e.g., GCN-Max, GCN-Mean \cite{defferrard2016convolutional}) directly embed the whole graph to a graph-level vector and compute the similarity between vectors as the similarity of the corresponding graph pairs.
However, these methods lose much node-level information, thereby reducing effectiveness. \emph{Matching-models} (e.g. SimGNN \cite{bai2019simgnn}, GSimCNN \cite{bai2018convolutional}, GMN \cite{li2019graph}) embed each node into a low-dimension vector which encodes both its own feature information and its local connection relationship information, and design different pairwise interaction strategies to compute the graph similarity score. However, the pairwise node comparison process needs at least quadratic time with respect to the number of nodes so that the problem of computation cost remains on large graphs.

In this paper, we focus on large graph similarity computation and try to address the following two challenges:
\begin{itemize}
\item For large graphs, the embedding-models are difficult to learn one vector representing the various features of a graph, so that they lose some accuracy. On the other hand, although the matching-models consider the fine-grained features of a graph, they can not efficiently calculate the similarity between graphs with many nodes. Thus, this challenge is \emph{how to achieve the trade-off between the accuracy and the efficiency of computing large graph similarity?}
\item Comparing with the graphs with fewer nodes, graphs with many nodes always contain distinct local features. For example, protein molecules are composed of many amino acids, determining the different functions of protein molecules. At the same time, amino acids are also composed of several atoms. As shown in Fig \ref{fig:1}, the subgraph in the red dashed box can be compared to an amino acid containing several atoms. When analyzing a protein molecule's function, only focusing on the whole protein molecule (the graph-level embedding) or only considering the atoms (the node-level embedding) may lose these local structural features. Thus, this challenge is \emph{how to learn embeddings capturing the local structure of large graph?}
\end{itemize}

To solve these challenges, we propose an end-to-end model PSimGNN, i.e., \emph{\underline{P}}artition based \emph{\underline{Sim}}ilarity Computation via \emph{\underline{G}}raph \emph{\underline{N}}eural \emph{\underline{N}}etworks.
First, the proposed model partitions each of the input graph into a set of subgraphs. To extract these local structural features, we design a novel graph neural network with an attention mechanism to map every subgraph into a subgraph-level embedding vector.
Next, we design an information interaction architecture with two modules to balance the subgraph-level and node-level information.
The first module conducts the coarse-grained similarity computation by
computing the similarity of these subgraph-level embedding vectors.
The second module conducts the fine-grained similarity computation by automatically selecting some of the subgraph pairs with higher similarity for node-level comparison, thus supplementing the subgraph-level embedding with fine-grained information.
Finally, the model integrates coarse-grained interaction information between subgraphs and fine-grained comparison information between nodes in different subgraph pairs to predict the final similarity score.
We evaluate the effectiveness of our model on large graph similarity computation. The experimental results show that PSimGNN outperforms the state-of-the-art graph similarity computation models. To summarize, our major contributions are:

\begin{itemize}
\item We first propose the graph partitioning based framework to address the challenging problem of similarity computation between large graphs. This framework achieves a good trade-off between accuracy and efficiency.
\item We propose a novel model that effectively extracts and aggregates local information to conduct a subgraph-level comparison. This can resolve the under-representation ability or high computation cost of many graph deep learning-based similarity computation models.
\item We conduct extensive experiments on a prevalent graph similarity/distance metric, GED, based on different size datasets. These experiments and theoretical analysis demonstrate the effectiveness and efficiency of PSimGNN model in graph similarity computation tasks.
\end{itemize}

The rest of this paper is organized as follows. In Section \ref{cha:related work}, we discuss the related works. In Section \ref{cha:PSimGNN}, we describe our model PSimGNN for graph similarity computation in detail and analyze the computation cost in theory. In Section \ref{others}, we compare the proposed model with some existing graph similarity computation methods on three datasets, introduce the experiment settings, and present the experimental results. In Section \ref{cha:conclusion}, we offer in-depth discussions and conclusions, and point out future directions.

\section{RELATED WORK}
\label{cha:related work}
In this section, we introduce the related works about graph partitioning, graph neural networks, graph similarity metrics and graph similarity computation.

\subsection{Graph Partitioning}
Graph partitioning is a way of cutting a graph into smaller pieces, while the nodes of these pieces are mutually exclusive with each other. Graph partitioning is an effective way for complexity reduction or parallelization \cite{bulucc2016recent} and the partitioned graph may be better suited for analysis and problem-solving than the original \cite{kaburlasos2009fuzzy}.
With the advent of ever-larger instances in applications such as scientific simulation, social networks, or road networks, graph partitioning, therefore, becomes more and more critical, multifaceted, and challenging \cite{bulucc2016recent}. Since graph partitioning is a hard problem, a variety of techniques and solutions are proposed. Global algorithms work
with the entire graph and compute a solution directly. These solutions can be improved using several heuristics, and high-quality graph partitioning solvers improve starting solutions. The most successful heuristic for partitioning large graphs is the multilevel graph partitioning approach. It consists of three phases: coarsening, initial partitioning, and uncoarsening \cite{bulucc2016recent}.

\subsection{Graph Neural Networks}
Graph Neural Networks (GNNs) is a useful framework for representation learning of graphs, directly operating on the graph structure. GNNs follow a neighborhood aggregation scheme, where the representation vector of a node is computed by recursively aggregating and transforming representation vectors of its neighboring nodes. Many GNN variants have been proposed and have achieved state-of-the-art results on both node and graph classification tasks \cite{ma2018graph, kazienko2012label}. Despite GNNs revolutionizing graph representation learning, there is a limited understanding of their representational properties.
Studies \cite{xu2018powerful} have shown that popular GNN variants (such as graph convolutional networks and GraphSAGE \cite{hamilton2017inductive}) have limited discriminative power, and they cannot learn to distinguish certain simple graph structures. In this paper, we use Graph Isomorphism Network (GIN) \cite{xu2018powerful} since GIN has been proven to be theoretically the most powerful GNN under the neighbor aggregation framework \cite{xu2018powerful}.

\subsection{GED \& MCS}
Graph Edit Distance (GED) \cite{bunke1983distance} can be considered as an extension of the String Edit Distance \cite{levenshtein1966binary} metric, which is defined as the minimum cost required to convert one graph to another through a sequence graph editing operations. Maximum Common Subgraph (MCS) \cite{bunke1998graph} is equivalent to GED under the same cost function \cite{bunke1997relation}. Both are the most common ways to calculate the similarity of graphs or the distance between graphs, which is the core operation of graph similarity search and many applications. However, this core operation, computing the GED or MCS between two graphs, is known to be NP-complete \cite{bunke1998graph,zeng2009comparing}. For a pair of graphs with more than 16 nodes, even the state-of-the-art algorithms cannot reliably compute the exact GED within reasonable time \cite{blumenthal2018exact}. So, instead of calculating the exact similarity, some methods can find approximate values in a fast and heuristic way. However, these methods usually require complicated design and the computation cost is still sub-exponential or polynomial in the number of nodes in the graphs, such as \textit{Hungarian} \cite{kuhn1955hungarian,riesen2009approximate}, \textit{VJ} \cite{fankhauser2011speeding,jonker1987shortest}, \textit{Beam}\cite{neuhaus2006fast}, etc.

\subsection{Graph Similarity Computation}
Computing the similarity of graphs is a basic and essential operation in many applications, including graph classification and clustering \cite{neuhaus2006edit}, social group network similarity identification \cite{steinhaeuser2008community} \cite{ogaard2013discovering}, object recognition in computer vision \cite{conte2004thirty}, and biological molecular similarity search \cite{kriegel2004similarity} \cite{tian2007saga}, etc. Graph similarity computation for metrics such as Graph Edit Distance (GED) is typically NP-hard, and existing heuristics-based algorithms usually achieve an unsatisfactory trade-off between accuracy and efficiency. Compared with traditional algorithms, which typically involve knowledge and heuristics specific to a metric, the neural network approaches learn graph similarity from data. During training, the parameters are learned by minimizing the loss between the predicted similarity scores and the ground truth; during testing, unseen pairs of graphs can be fed into these models for fast approximation of their similarities \cite{bai2020learning}.

This paper is the first attempt towards large-scale graph similarity computation with deep learning methods.
Current deep learning methods for graph similarity computation can be classified as embedding models and matching models. Embedding models such as GCN-Mean and GCN-Max, directly map each graph to a feature vector and compute the similarity score between these feature vectors. Embedding models are efficient, but the performance is usually low due to the lack of interactions across graphs. Matching models, including GMN, SimGNN, and GSimCNN, embed a pair of graphs at the same time with a cross-graph matching mechanism. They are more accurate, but the cross-graph matching process often brings a significant increase in time consumption (at least quadratic computation cost over the number of nodes). Thus, our work explores how to compute the similarity between large-scale graphs while maintaining high accuracy efficiently.

\begin{figure*}
\centering
\includegraphics[width=\textwidth]{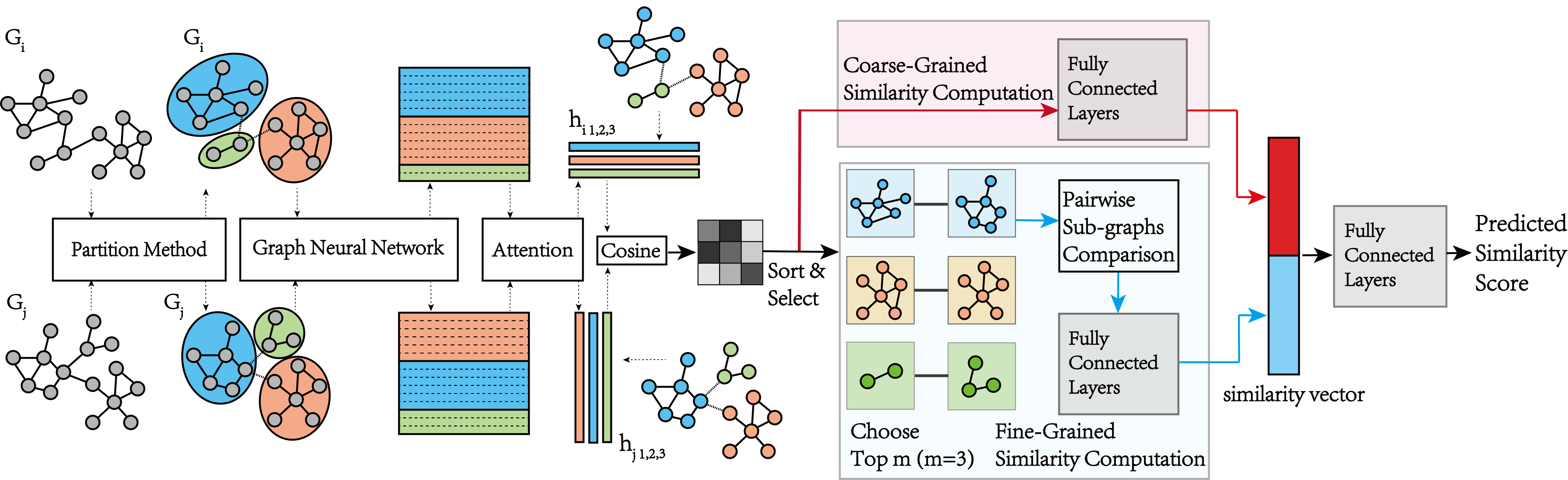}
\caption{The general architecture of our model. The red arrows denote the data flow for subgraph-level interaction and the blue arrows denote the data flow for node-level comparison. After the graph partitioning method, only the top $\bm{m}$ (here $\bm{m = 3}$) subgraph pairs with the highest similarity scores will conduct the node-level comparison.}
\label{fig:general_architecture}
\vspace{-4mm}
\end{figure*}

\section{The Proposed Approach: PSimGNN}
\label{cha:PSimGNN}

In this section, we formally define the problem of graph similarity computation, and then  introduce the proposed method PSimGNN, i.e., \emph{\underline{P}}artition based \emph{\underline{Sim}}ilarity Computation via \emph{\underline{G}}raph \emph{\underline{N}}eural \emph{\underline{N}}etworks, which is an end-to-end neural network-based method to solve graph similarity computation problem. PSimGNN consists of four parts: (1) graph partitioning; (2) subgraph-level embedding interaction; (3) node-level comparison; (4) graph similarity score computation. An overview of PSimGNN is shown in the Figure~\ref{fig:general_architecture}.

\subsection{Problem Definition}
We define an undirected and unweighted graph $G=\{V,E\}$, where $V=\{v_1,...,v_{|V|}\}$ is a set of nodes and $E=\{e_1,...,e_{|E|}\}$ is a set of edges.
$H \in \mathbb{R}^{N \times D}$ represents node features, where $N$ is the number of nodes in graph $G$ (or $N$ = $|V|$) and $D$ is the dimension of node feature vectors.
We transform GED into a similarity metric ranging between 0 and 1.
Our goal is to learn a neural network-based function that takes two graphs as input and outputs the similarity score that can be transformed back to GED through a one-to-one mapping.

\begin{figure*}
\includegraphics[width=\textwidth]{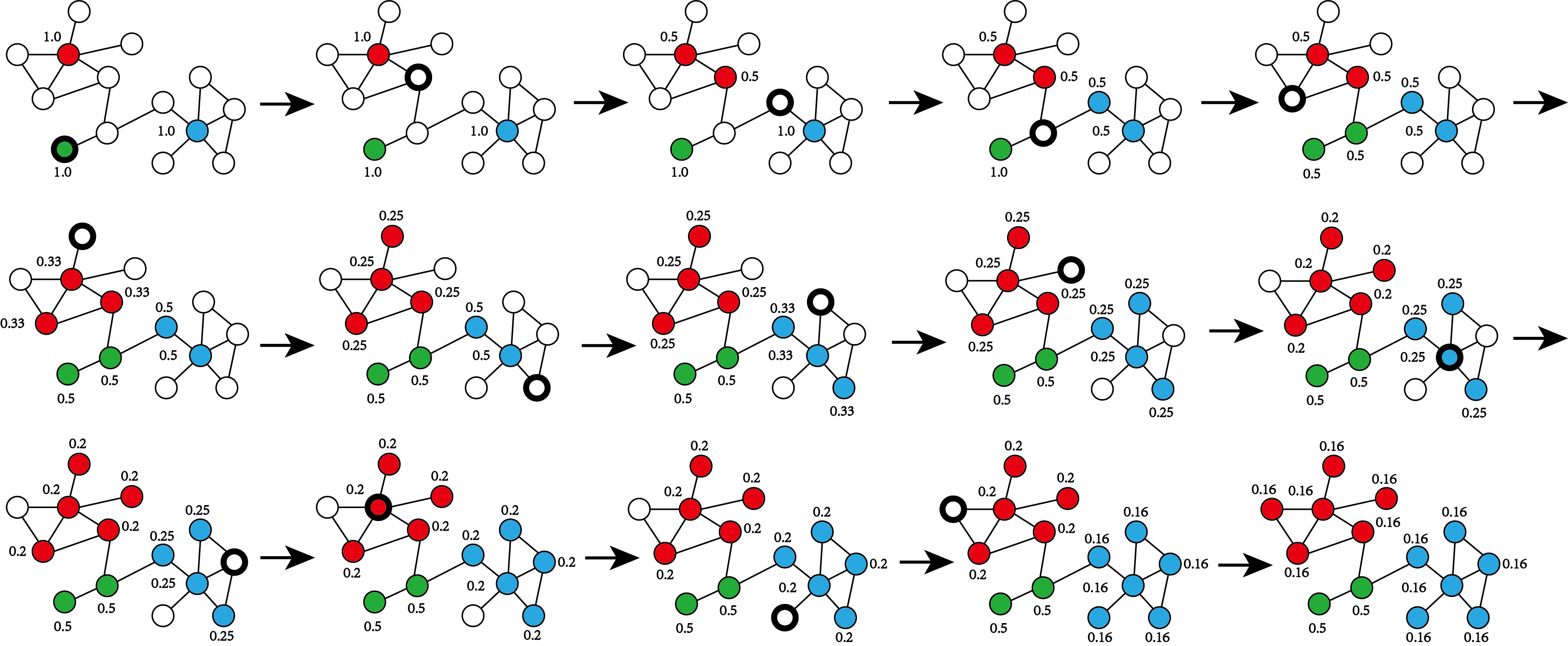}
\caption{The workflow of $\bm{FluidC}$ for $\bm{k_f = 3}$ communities (red, green and blue). Each node assigned to a community is labeled with the density of that community. The update rule is evaluated on each step for the node highlighted in black. Only the schematic results of the first iteration are given here.}
\label{fig:FluidC}
\vspace{-3mm}
\end{figure*}

\subsection{Graph Partitioning}

Most neural network-based graph similarity computation models use appropriate mechanisms to generate graph-level embeddings and node-level embeddings and calculate the graph similarity score between different graphs combining a coarsen-grained graph-level interaction and a fine-grained node-level comparison. However, for graphs with a large number of nodes, these approaches may have several limitations:
\begin{itemize}
\item Only graph-level embedding may have limited ability to represent the whole graph. Sometimes we have to pay attention to some local structure characteristics.
\item Due to a large number of nodes, the node-level comparison will bring high computation cost, and too much matching between nodes far away will also introduce some noise.
\end{itemize}

To overcome these limitations and better reflect large graphs' local structure characteristics, we partition a graph into $k_f$ subgraphs using the graph partitioning method. In our experiments, the Fluid Communities algorithm $(FluidC)$ \cite{pares2017fluid} shows the best performance. Graph partitioning contains three steps:
\begin{itemize}
\item Step-1: Choose $k_f$ nodes randomly in the graph as the initial nodes of $k_f$ communities.
\item Step-2: Iterate over all nodes in random order and update each node's community based on its community and the communities of its neighbors.
\item Step-3: Repeat step-2 until convergence.
\end{itemize}

Here we focus on its updating strategy, or more precisely, step 2.
For a graph $G = (V, E)$ composed of a set of vertices $V$ and a set of edges $E$, FluidC initializes $k_f$ fluid communities $\mathcal{C} = \{c_1, c_2, ..., c_{k_f}\}$, where $0<k_f\leq|V|$. Each community $c\in \mathcal{C}$ is initialized in a different and random vertex $v\in V$. Each initialized community has an associated density $d$ in the range (0,1]. More precisely, the density of a community is the inverse of the number of vertices that make up the community: $d(c) = 1/|v\in c|$. We can notice that a fluid community composed of a single vertex (for example, each community at initialization) has the largest possible density ($d = 1.0$).

The algorithm traverses all $V$ in random order and uses the updating rule to update the community to which each vertex belongs. When the vertex distribution to the community has not changed in two consecutive steps, the algorithm has converged and ended. Next, we focus on its updating rules.

The updating rule for a specific vertex $v$ returns the community or communities with maximum aggregated density within the ego network of $v$. The updating rule is formally defined in equations \ref{update:1} and \ref{update:2}.
\begin{equation}
    \mathcal{C}_v' = argmax_{c\in \mathcal{C}} \sum_{w\in \{v,\Gamma(v)\}} d(c) \times \delta(c(w), c)
\label{update:1}
\end{equation}
\begin{equation}
    \delta(c(w), c) = \left\{
             \begin{array}{lr}
             1,\ if\ c(w)=c &  \\
             0,\ if\ c(w)\neq c& \\
             \end{array}
\right.
\label{update:2}
\end{equation}
where $v$ is the vertex which is going to be updated, $\mathcal{C}_v'$ is the set of candidates to be the new community of $v$, $\Gamma (v)$ are the neighbours of $v$, $d(c)$ is the density of community $c$, $c(w)$ is the community vertex $w$ belongs to and $\delta(c(w), c)$ is the Kronecker delta.

In some exceptional cases, $\mathcal{C}_v'$ can contain multiple community candidates with equal maximum sum. If $\mathcal{C}_v'$ has the current community of vertex $v$, then $v$ will not change its community. However, if $\mathcal{C}_v'$ does not contain the current community of $v$, the update rule will select a random community in $\mathcal{C}_v'$ as the new community of $v$. Here we give a formal representation of the updating rule:
\begin{equation}
    c'(v) = \left\{
             \begin{array}{lr}
             x\sim \mathcal{S}(\mathcal{C}_v'),\  & if\ c(v)\notin \mathcal{C}_v'  \\
             c(v),\ & if\ c(v)\in \mathcal{C}_v' 
             \end{array}
\right.
\label{update:3}
\end{equation}
where $c'(v)$ is the community of the vertex $v$ of the next step, $\mathcal{C}_v'$ is the set of candidate communities, and $x\sim \mathcal{S}(\mathcal{C}_v')$ is the random sampling from the discrete uniformly distribution of $\mathcal{C}_v'$.

The entire process can refer to Figure \ref{fig:FluidC}. At all times, each community has a total density of 1, which is equally distributed among the nodes it contains. If a node changes its community, node densities of affected communities are adjusted immediately. When a complete iteration over all nodes is done, such that no node changes the community it belongs to, the algorithm has converged and returns.
Through $FluidC$, we can obtain a series of connected subgraphs (or communities) that can reflect local features. The similarity computation at the subgraph-level and node-level can be performed later.

\subsection{Subgraph-level Comparison}

One useful graph-level embedding can efficiently preserve the structural information, and the similarity between two graphs can be computed by interacting with the two graph-level embeddings. For graphs with many nodes, by comparing the similarity between different subgraphs generated by some graph partitioning methods (like $FluidC$ in our experiment), the local similarity between two large graphs can be better reflected. The entire process involves the following three parts: (1) \textbf{Subgraph node embedding}, which embeds the nodes of each subgraph into vectors, encoding its structural information; (2) \textbf{Subgraph embedding}, which embeds each subgraph into one graph-level vector considering the context information through an attention-based node aggregation way; (3) \textbf{Subgraph-subgraph interaction}, which receives two subgraph-level embeddings and returns the interaction score representing the similarity between subgraphs. Next, these subgraph interaction scores are further reduced to a final similarity score through Multilayer Perceptron, representing the similarity of the pair of large graphs. And the parameters involved in these three steps can be updated by comparing the final similarity score with the ground truth similarity score in the training process.

\subsubsection{Part I: Subgraph Node Embedding}

Among the existing multiple graph neural network methods, we choose Graph Isomorphism Network (GIN) \cite{xu2018powerful} because it can not only efficiently gather information of neighboring nodes like Graph Convolutional Networks (GCN) \cite{kipf2016semi, defferrard2016convolutional} and GraphSAGE \cite{hamilton2017inductive}, but also learn accurate structural information. The hidden layer can be written as follow:
\begin{equation}
h_v^{(k)} = MLP^{(k)}\big((1+\epsilon^{(k)})\cdot h_v^{(k-1)}+\sum_{u\in \mathcal N(v)}h_u^{(k-1)}\big)
\end{equation}
, where $h_v^{(k)}$ is the $k$-th layer node embedding for the node $v$, $MLP$ means Multilayer Perceptron \cite{pal1992multilayer}, $\epsilon$ is a learnable parameter and $\mathcal N(v)$ represents the neighbor nodes of node $v$.

We treat each node as the same label for graphs with unlabeled nodes, thereby obtaining the same constant as the initial representation. After multiple GIN layers (3 layers in our experiment), the node-level embeddings information will be fed into the attention module as described below.

\subsubsection{Part II: Subgraph Embedding}

This model uses a weighted sum method, where we use an attention mechanism to generate subgraph-level embeddings with a weighted sum method. Instead of averaging all nodes or giving each node different weights according to the node's degree, our attention module focuses more on the nodes that can better represent the full graph structure information.

After learning the node-level embedding, the node embeddings in subgraph can be expressed as $X\in \mathbb{R}^{N_s\times D}$, where $N_s$ represents the number of nodes in the subgraph, and $D$ is the dimension of each node embedding. The representation of the whole subgraph information can be written as $\emph{z}\in \mathbb{R}^D$, which is a non-linear expression of the average value of $N$ nodes embedding: $z = {\rm tanh}((\frac{1}{N}\sum_{i=1}^{N}{u_i})W_z)$, where $W_z$ is a learnable weight matrix.

By learning the weight matrix, $z$ provides the subgraph's global structure and feature information suitable for a given similarity measure. Then based on $z$, we can calculate an attention weight for each node. In short, attention works as a memory-access mechanism by generating larger attentive coefficients for input features that are relevant to the learning task. In this paper, we adopt an attention mechanism to generate subgraph-level representation. For node $i$, to notify the global information, we take the inner product between its node embedding $x_i$ and $z$. That is to say, the nodes that are more capable of expressing the features of the graph should be given higher weights. The sigmoid function $\sigma(x) = 1 + exp^{-x}$ is applied to the result to ensure that the attention weight is between (0, 1). Finally, subgraph embedding $h\in \mathbb{R}^{D}$ is the weighted sum of node embedding:
\begin{equation}
h = \sum_{j=1}^{N}\sigma(x_j\odot z)x_j = \sum_{j=1}^{N}\sigma\Big(x_j\odot {\rm tanh}\big((\frac{1}{N}\sum_{i=1}^{N}{u_i})W_z\big)\Big)x_j
\end{equation}
, where $\odot$ represents the dot product between vectors.

\subsubsection{Part III: Subgraph-subgraph Interaction}

Through the node embedding and attention mechanism mentioned above, we have achieved subgraph-level embedding. Good node embedding and attention mechanisms should embed graphs with similar structures and similar features in similar positions in space, so their distance should be relatively small. Here we use the $cosine$ similarity to measure the similarity between a pair of subgraph embeddings:
\begin{equation}
s(h_1,h_2) = cos(h_1,h_2) = \frac{h_1\odot h_2}{||h_1||_2||h_2||_2}
\end{equation}
, where $||h||_2$ is the 2-norm of $h$.

A pair of large graphs are partitioned into $k$ subgraphs respectively, and the similarity between two subgraphs between large graphs is calculated using the method mentioned above. After that, $k^2$ similarity scores are obtained, and Multilayer Perceptron ($MLP$) is used to map these $k^2$ scores to the final similarity score to characterize the similarity between the pair of large graphs:
\begin{equation}
	s(G_1,G_2)=MLP\Big(\bigoplus_{i=1,j=1}^{k}s(G_1^i,G_2^j)\Big)
\end{equation}
, where $\bigoplus$ is the concatenation operation, $s(G_1,G_2)$ represents the similarity score between the pair of large graphs and $s(G_{1}^i,G_{2}^j)$ represents the similarity score between the $i$-th subgraph of $G_1$ and the $j$-th subgraph of $G_2$.

\subsection{Node-level Comparison}

\begin{figure}[t]
\centering
\includegraphics[width=0.9\linewidth]{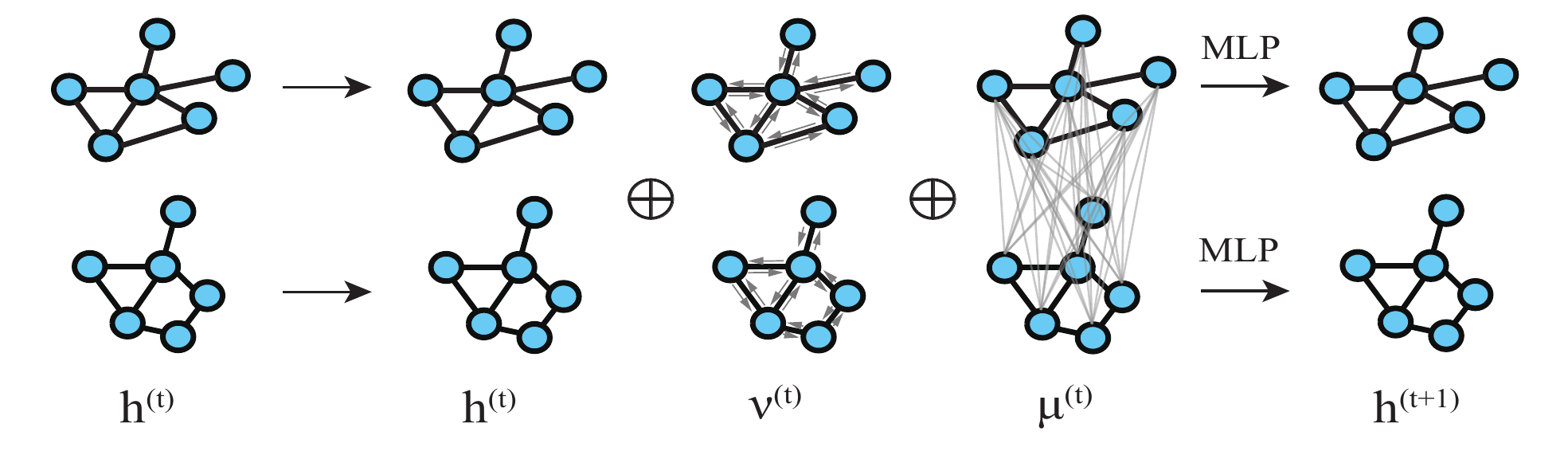}
\caption{Propagation layer in node-level comparison. Each round of iteration is based on the embeddings of the previous round and the node-level comparison within each graph and between graphs.}
\label{fig:Propagation layer in node level comparison}
\vspace{-4mm}
\end{figure}

Only considering subgraph-level embedding interaction may lose some fine-grained node-level information, so we design the following node-level comparison component utilizing the interaction between nodes in subgraphs.

This component accepts a pair of subgraphs as its input, and calculates the similarity between them through the comparison of nodes within each subgraph and between the pair of subgraphs. An overview of the interaction is shown in the Figure~\ref{fig:Propagation layer in node level comparison}, where $h^{(t)}$ represents the node embeddings in each subgraph after $k$-th propagation layer. We assume that the input subgraph pair can be represented as $G_1^m$, $G_2^n$, and their node sets and edge sets are $V_1^m$, $V_2^n$ and $E_1^m$, $E_2^n$, respectively. After $t$ iterations within the graph and between graphs, the embedding of node $i$  can be represented as $h_i^{(t)}$.
In each interaction within the subgraph, the influence of node $j$ on node $i$ is:
\begin{equation}
\nu_{j\to i} = MLP(h_i^{(t)}\oplus h_j^{(t)}),\forall (i,j) \in E_1^m\cup E_2^n
\end{equation}
Then is the interaction between subgraphs for cross-graph communication. An attention mechanism is used to give different weights to the nodes of another subgraph to indicate the importance of different nodes $j$ to nodes $i$:
\begin{equation}
a_{j\to i} = \frac{\rm{exp}(h_i^{(t)}\odot h_j^{(t)})}{\sum_{j^{'}}\rm{exp}(h_i^{(t)}\odot h_{j^{'}}^{(t)})}
\end{equation}

Through this attention mechanism, we magnify the influence between similar nodes in one pair of subgraphs, and use $\mu_{j\to i}$ to represent the interaction between node $j$ and node $i$ in different subgraphs:
\begin{equation}
\mu_{j\to i} = a_{j\to i}(h_i^{(t)}-h_j^{(t)}),\forall i\in V_1,j\in V_2, or \ \forall i\in V_2,j\in V_1
\end{equation}

After obtaining the interactive information within each subgraph and between one pair of subgraphs, we merge the $t$-th round propagation node information with it, and then generate the $(t+1)$-th round propagation node information:
\begin{equation}
h_i^{(t+1)} = MLP\Big(h_i^{(t)}\oplus \sum_{j}\nu_{j\to i} \oplus \sum_{j^{'}}\mu_{j^{'}\to i}\Big)
\end{equation}

After iterating through $T$ rounds, we get the embedding of each node, denoted as $h^{(T)}$, and then through a self-attention mechanism aggregation layer, we get a subgraph-level embedding:
\begin{equation}
h_{agg} = MLP_{agg}\Bigg(\sum_{i\in V}\sigma(MLP_{att}(h_i^{(T)}))\odot MLP(h_i^{(T)})\Bigg)
\end{equation}

After obtaining the fine-grained embedding of each subgraph, we use the $cosine$ similarity to measure the similarity between one pair of graphs, which is expressed as:
\begin{equation}
s(h_{agg1},h_{agg2}) = cos(h_{agg1},h_{agg2}) = \frac{h_{agg1}\odot h_{agg2}}{||h_{agg1}||_2||h_{agg2}||_2}
\end{equation}

\subsection{Graph Similarity Score Computation}

It is worth mentioning that through the previous graph partitioning, each large graph is partitioned into $k$ subgraphs, and there will be $k^2$ pairs of subgraphs. Here, we sort the subgraph-level similarities obtained before, and only the pairs with top $m$ similarity score will perform a node-level comparison. We use $(MLP)$ to integrate $k^2$ coarse-grained scores and $m$ fine-grained scores to finally get the similarity between the large graphs:
\begin{subequations}
	\begin{equation}
	s(G_1,G_2)_{coarse}=MLP\Big(\bigoplus_{i=1,j=1}^{k}s(G_1^i,G_2^j)\Big)
	\end{equation}
	\begin{equation}
	s(G_1,G_2)_{fine}=MLP\Big(\bigoplus_{t=1}^{m}s(G_1^{it},G_2^{jt})\Big)
	\end{equation}
	\begin{equation}
	s(G_1,G_2) = MLP\Big(s(G_1,G_2)_{coarse}\oplus s(G_1,G_2)_{fine}\Big)
	\end{equation}
\end{subequations}

After the similarity score, $s(G_1,G_2)\in \mathbb{R}$, is predicted, it is compared with the ground truth similarity score, $s(G_1,G_2)_{gt}\in \mathbb{R}$, using the following mean square error loss function:
\begin{equation}
\mathcal{L} = \frac{1}{\mathcal{|T|}}\sum_{(i,j)\in\mathcal{T}}\Big(s(G_1,G_2)-s(G_1,G_2)_{gt}\Big)^2
\end{equation}
,where $\mathcal{T}$ is the training graph pairs and $\mathcal{|T|}$ is the total number of the training graph pairs.

\subsection{Efficiency Analysis}

For a pair of input graphs $G_1$ and $G_2$ with $E_1$, $E_2$ edges and $N_1$ and $N_2$ nodes separately, we can evaluate the efficiency of several types of models that are commonly used in graph similarity computation.

Then we analyze the efficiency of PSimGNN and discuss how it can improve the efficiency by graph partitioning.
Note that there exists a lot of variance for each model. We only use the simplest cases.

\subsubsection{Embedding models}
The embedding model refers to calculating the similarity between graphs by generating graph-level embeddings. Assuming the simplest case here, we only visit every edge once and deploy two computational operations on the two nodes it connects, contributing to the feature of local topology. Thus the computation cost for these cases is $O\left(max(E_1,E_2)\right)$.

\subsubsection{Matching models}
The matching model refers to calculating the similarity between graphs by matching (graph-level interaction or node-level comparison).
Assuming the simplest case here, we compute the relationship across $N_1$ and $N_2$. This part involves $N_1\times N_2$ computational operations because we have to calculate the connection between every node in $G_1$ to all nodes in $G_2$. For the common matching models, both SimGNN and GSimCNN pad fake nodes to the smaller graph at the node-level comparison to emphasize their size difference. GMN also has the interaction of nodes within each graph, so the final computation cost is $O(max(N_1, N_2)^2)$.

\subsubsection{PSimGNN}
The computation cost of PSimGNN can be divided into three parts to analyze. 
(1) Graph Partitioning. 
In our model, we choose $FluidC$ as the graph partitioning method. 
As analyzed in \hyperref[subsec: Graph Partitioning]{Section 3.1}, it updates node information based on neighbor nodes or the connected edges of nodes, so it belongs to the fastest and most scalable family of algorithms in the literature with a linear computation cost of $O(E)$ \cite{pares2017fluid}. 
Notice that the partitioned subgraphs can be pre-computed and stored. In the setting of graph similarity search, the unseen query graph only needs to be partitioned once to obtain its subgraphs. 
(2) Subgraph-level Embedding Interaction. 
The computation cost associated with the generation of node-level and subgraph-level embeddings is $O(E)$\cite{kipf2016semi}. 
Assuming that each graph is partitioned into $k$ subgraphs and the embedding dimension at the subgraph-level is $D$, we use $cosine$ to measure the similarity between embeddings. The computation cost in the subgraph interaction part is $O(Dk^2)$. 
As mentioned above, the sub-graph level and node-level embedding can also be saved in advance, which significantly saves graph similarity query cost. 
(3) Node-level Comparison. 
According to the $k^2$ similarity scores obtained by subgraph-level interaction, we select top $m$ subgraph pairs with the highest similarity scores for node-level comparison. 
After partitioning, the average number of nodes in each subgraph is $N_1/k$ or $N_2/k$. 
As analyzed in \hyperref[subsec: Matching models]{Section 3.5.2}, the average node-level comparison computation cost of one pair of subgraphs is $O(max(N_1/k,N_2/k)^2)$. 
Since we choose $m$ pairs, the total computation cost of this part is $O(m\times max(N_1/k,N_2/k)^2)$ or $O(m/k^2\times max(N_1,N_2)^2)$, where the range of $m$ belongs to $\{0, 1, 2, ..., k^2\}$.
The parameter $m$ can be used as a hyperparameter to adjust the relationship between accuracy and time. 
When $m$ is set to zero, our model only calculates the coarse-grained subgraph similarity. 
At this time, the model's computation cost is $O(E)$, where $E$ is the number of edges in the large graph. 
When $m$ is set to $k^2$, our model performs fine-grained similarity calculation for each pair of subgraphs, and the computation cost of the model is $O(N^2)$, where $N$ is the number of nodes in the large graph.
For occasions with time requirements, we can only perform coarse-grained matching between subgraphs. For occasions where accuracy requirements are relatively high, we can perform a fine-grained node-level comparison to improve model performance.
Therefore, according to specific application scenarios, trade-offs between time and accuracy can be made to choose the best solution.

\subsubsection{Similarity search}
In the similarity search problem, we assume that we have a database consisted of $K$ graphs, each of which has $N$ nodes and $E$ edges, for simplicity. We need to finish computing the similarity between all the graphs in the database and an incoming new graph (also with $N$ nodes and $E$ edges).
In embedding models, we can compute all the feature vectors for graphs in the database at the very beginning.
And then, when the new graph comes, we encode it to its feature vector and only compute similarity based on the feature vectors.
Thus the computation cost is $O(E\times K)$.
We can only forward pairs of graphs in matching models every time because of the computation across graphs. Thus the computation cost is extremely high $O(N^2\times K)$.
And obviously, the computation cost for our framework is $O(m/k^2 \times N^2 \times K + E \times K)$.
When $m$ is small, the computation cost becomes $O(E\times K)$; when $m$ is large, the computation cost becomes $O(m/k^2 \times N^2 \times K)$.
This also reflects the adjustability of our model.
It is worth mentioning that our model is not suitable for very dense graphs because it is challenging to get subgraphs that can better reflect local information.
In our discussion, $E << N^2$.

\begin{table*}[t]
\renewcommand\arraystretch{1.2}
\caption{Statistics of datasets.}
\label{dataset}
\centering
\scalebox{0.8}{
\begin{tabular}{|c|ccccccccc|}
\hline
Dataset &Graph Meaning &\#Graphs &\#Pairs &\multicolumn{1}{m{1cm}<{\centering}}{Min \#Nodes} &\multicolumn{1}{m{1cm}<{\centering}}{Max \#Nodes} &\multicolumn{1}{m{1cm}<{\centering}}{Avg \#Nodes} &\multicolumn{1}{m{1cm}<{\centering}}{Min \#Edges} &\multicolumn{1}{m{1cm}<{\centering}}{Max \#Edges} &\multicolumn{1}{m{1cm}<{\centering}|}{Avg \#Edges}
\\ \hline
AIDS &Chemical Compounds &700 &490K &2 &10 &8.90 &1 &14 &8.80\\
LINUX  &Program Dependency Graphs &1000 &1M &4 &10 &7.58 &3 &13 &6.94\\
IMDB  &Actor/Actress Ego-Networks &1500 &2.25M &7 &89 &13.00 &12 &1467 &65.95\\\hline
BA-60 &Barabási–Albert graph with 60 nodes &200 &40K &54 &65 &59.50 &54 &66 &60.06\\
BA-100 &Barabási–Albert graph with 100 nodes &200 &40K &96 &105 &100.01 &96 &107 &100.56\\
BA-200 &Barabási–Albert graph with 200 nodes &200 &40K &192 &205 &199.63 &193 &206 &200.16\\
IMDB-X & Filtered Actor/Actress Ego-Networks &220 &48.4K &15 &52 &21.35 &33 &186 &74.20\\\hline
\end{tabular}}
\vspace{-3mm}
\end{table*}

\begin{figure*}[t]
\centering
\subfigure[BA-60]{
\begin{minipage}[t]{0.4\linewidth}
\centering
\includegraphics[width=1\linewidth]{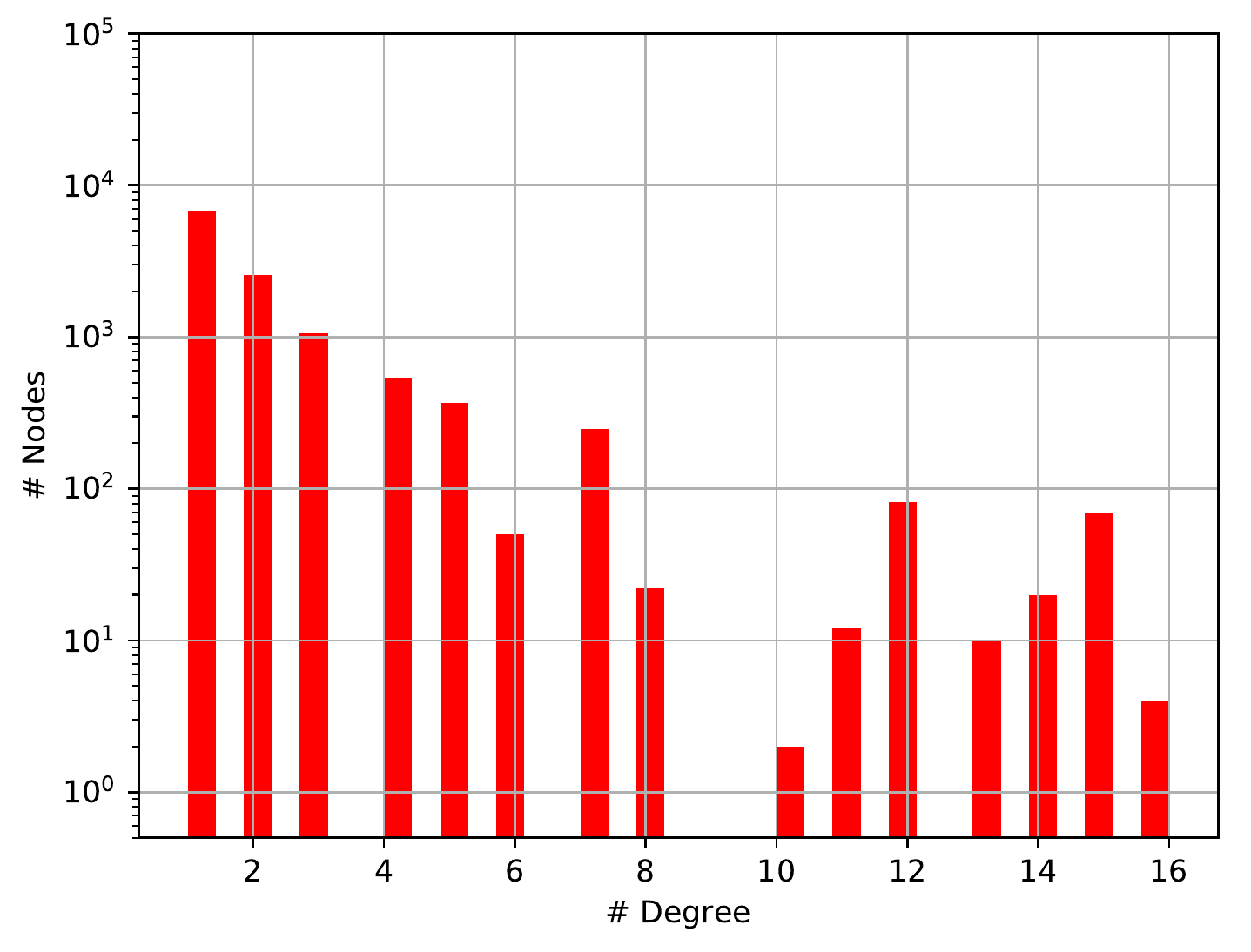}
\end{minipage}
}
\subfigure[BA-100]{
\begin{minipage}[t]{0.4\linewidth}
\centering
\includegraphics[width=1\linewidth]{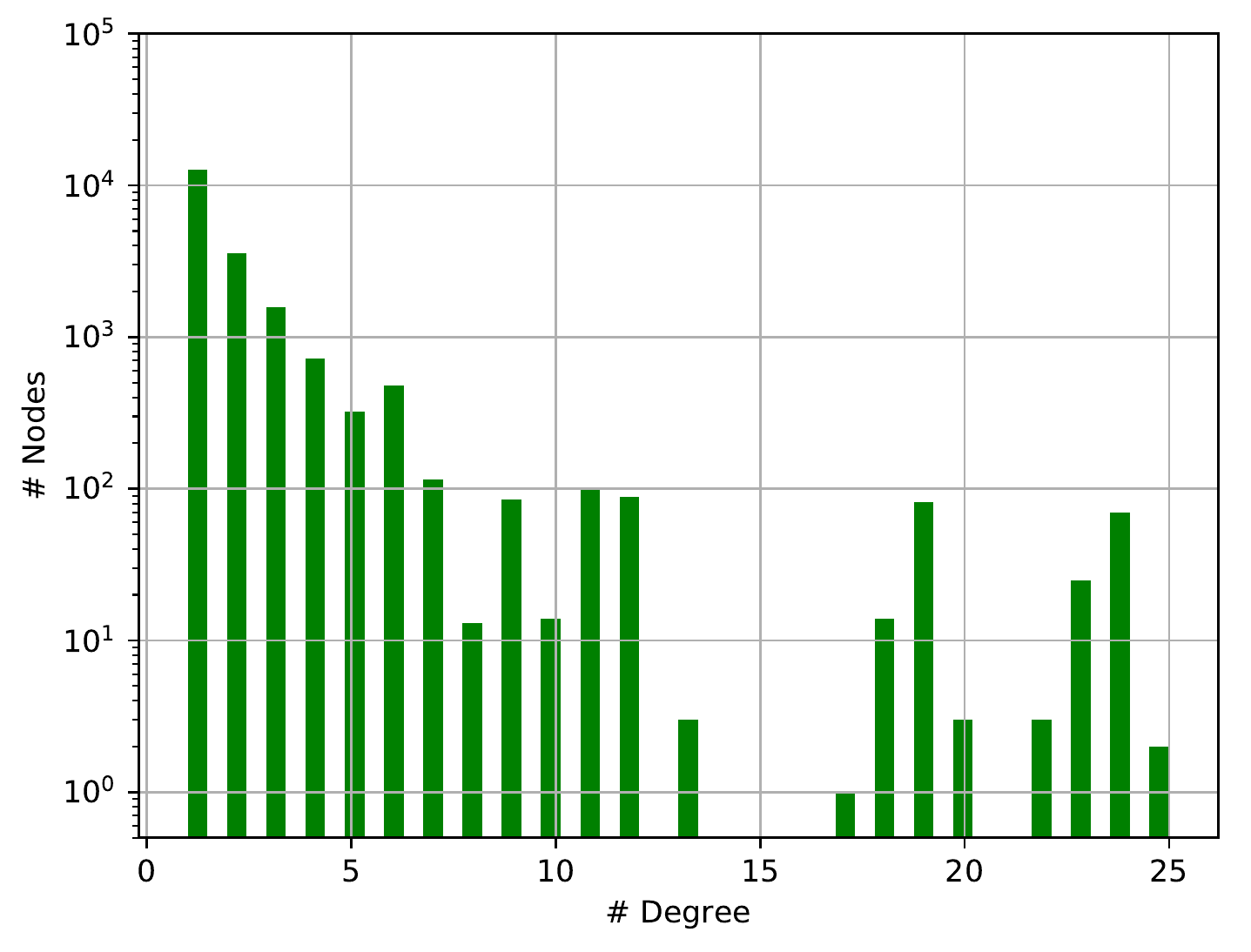}
\end{minipage}
}
\\
\subfigure[BA-200]{
\begin{minipage}[t]{0.4\linewidth}
\centering
\includegraphics[width=1\linewidth]{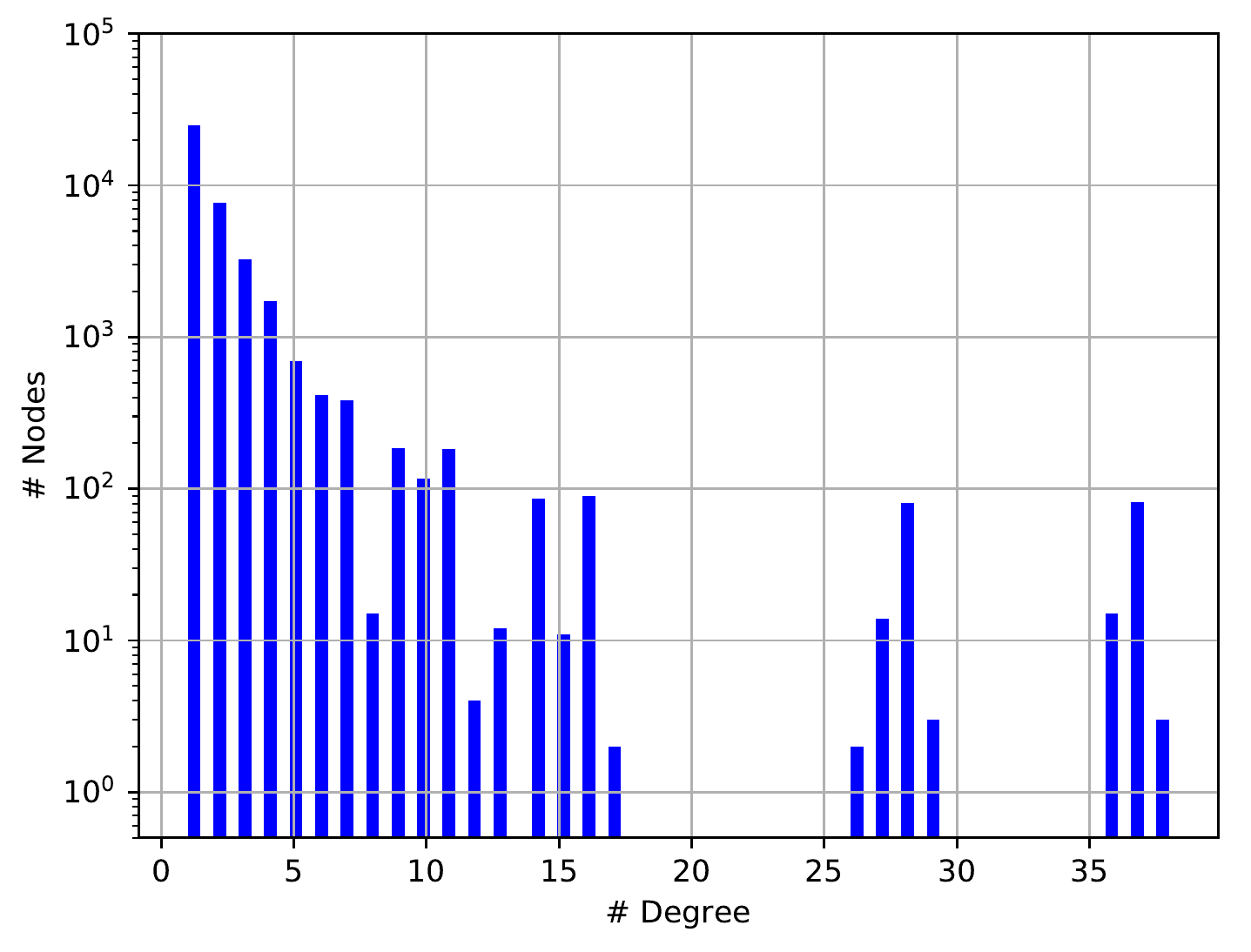}
\end{minipage}
}
\subfigure[IMDB-X]{
\begin{minipage}[t]{0.4\linewidth}
\centering
\includegraphics[width=1\linewidth]{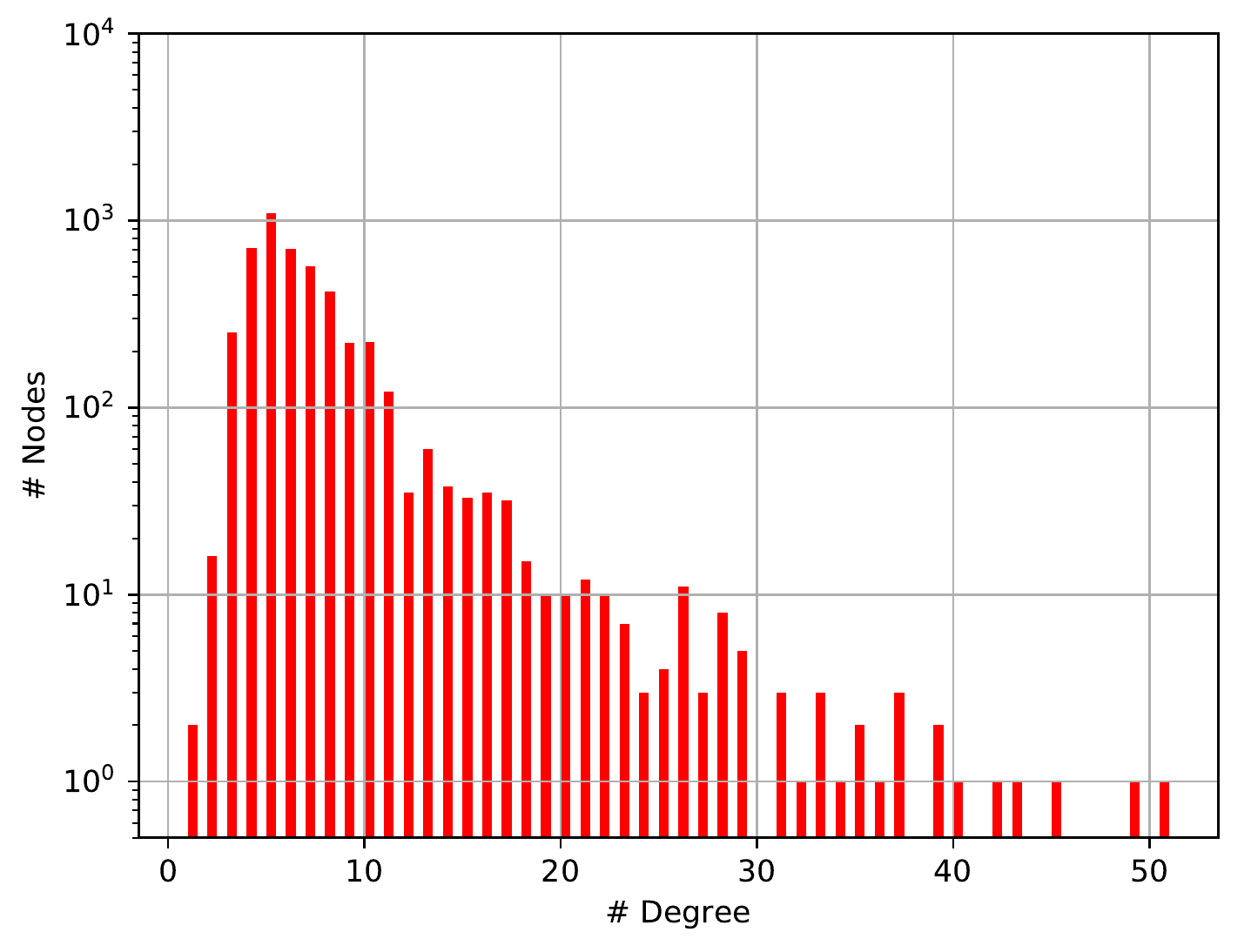}
\end{minipage}
}
\centering
\caption{Nodes degree distribution of BA-model datasets.}
\label{fig:Degree}
\vspace{-5mm}
\end{figure*}

\section{Experiments}
\label{others}

\subsection{Datasets}

In this section, we first introduce a graph similarity computation dataset based on Barab ási–Albert preferential attachment model (BA-model) \cite{jeong2003measuring}, which consists of three sub-datasets: BA-60, BA-100, BA-200, named according to the average number of nodes per graph.
A real-world dataset is also introduced to better demonstrate our framework's effectiveness: The Internet Movie Database (IMDB)\footnote{http://www.imdb.com/interfaces\#plain}.
The details about these datasets are as following, and we compare these with other well-known datasets used for graph similarity computation.

\subsubsection{Barabási–Albert model dataset}
Here we introduce the Barabási–Albert model (BA-model) concept, the rules for generating a Barabási–Albert graph (BA-graph), and how our datasets are produced. The BA-model \cite{jeong2003measuring} is an algorithm for generating random scale-free networks using a preferential attachment mechanism. Several natural and human-made systems, including the Internet, the world wide web, citation networks, and some social networks, are thought to be approximately scale-free and contain few nodes (called hubs) with unusually high degrees compared to the other nodes of the network. The BA-model tries to explain such nodes in real networks and incorporates two important general concepts: growth and preferential attachment, which exist widely in real networks. Growth means that the number of nodes in the network increases over time and preferential attachment indicates that the more connected a node is, the more likely it is to receive new links. Nodes with a higher degree have a more vital ability to grab links added to the network.

The BA-model begins with an initial connected network of $m_0$ nodes.
New nodes are added to the network one at a time. Each new node is connected to $m\leq m_{0}$ existing nodes with a probability proportional to the number of links that the existing nodes already have. Formally, the probability $p_i$ that the new node is connected to node $i$ is $p_i=\frac{k_i}{\sum_jk_j}$ \cite{albert2002statistical}, where $k_i$ is the degree of node $i$ and the sum is made overall pre-existing nodes $j$ (i.e. the denominator results in twice the current number of edges in the network). Heavily linked nodes ("hubs") tend to quickly accumulate even more links, while nodes with only a few links are unlikely to be chosen as the destination for a new link. The new nodes have a "preference" to attach themselves to the already heavily linked nodes.

Our datasets are made up of some basic graphs and derivative graphs that have been trimmed, which solve several problems:

\begin{itemize}
\item When generating a graph with a large number of nodes randomly, there is a high probability that the generated graphs are dissimilar between each other, which results in an uneven similarity distribution after normalization.
\item Due to a large number of nodes in each graph, the approximate GED algorithm cannot guarantee that the calculated similarity can fully reflect the graph pairs' similarity. We trim and generate derivative graphs while recording the number of trimming steps. These steps and the values calculated by the approximation algorithm take the minimum value as the GED with the basic graph, thereby obtaining a more accurate similarity.
\item By trimming different steps, we can generate graphs with different similarities, which is more conducive to the experiment of graph similarity query.
\end{itemize}

There are three trimming methods: delete a leaf node, add a node, and add an edge. Since deleting an edge may have a greater impact on the generated graph, we will not consider this method. We try to trim the base graph without changing the base graph's global features to generate more similar graph pairs. In this way, we get three datasets according to the following generation rule.

A BA-graph of $n$ nodes is grown by attaching new nodes, each with $m$ edges that are preferentially attached to existing nodes with a high degree. We set $n$ to be 60, 100, and 200, respectively, and $m$ is fixed to 1 to generate basic graphs. Each sub-dataset generates two basic graphs, and each base graph is trimmed with different GEDs. For each basic graph, generate 99 trimmed graphs in the range of GED 1 to 10. So each sub-dataset consists of two basic graphs and 198 trimmed graphs.

\subsubsection{The Internet Movie Database}
The Internet Movie Database (IMDB) consists of the entities of movies, actors, producers, and their relationships.
We filter the original IMDB dataset based on two principles: (1) graphs that have 15 or more nodes, (2) graphs where the ratio of the number of edges to the number of nodes is less than 5.
The graphs filtered in this way have a sufficient number of nodes, and they are not too dense, which are suitable for partitioning in the task of graph similarity computation.
In this paper, we call the new dataset IMDB-X.

\subsubsection{Comparison with Other Datasets}

Because in other public datasets, such as AIDS \cite{liang2017similarity} and LINUX \cite{wang2012efficient}, the number of nodes in each graph is relatively small and local structures are not obvious, the characteristics of the entire graph can be easily represented.
As for IMDB \cite{yanardag2015deep}, (named "IMDB-MULTI") there are some graphs with a large number of nodes. However, these graphs are relatively dense, and too many edges between nodes will make the local structures less obvious.
So we filter the IMDB as IMDB-X and focus on the BA datasets and IMDB-X.

\begin{figure}[t]
\centering
\subfigure[BA60-20604.gexf]{
\begin{minipage}[t]{0.4\linewidth}
\centering
\includegraphics[width=1\linewidth]{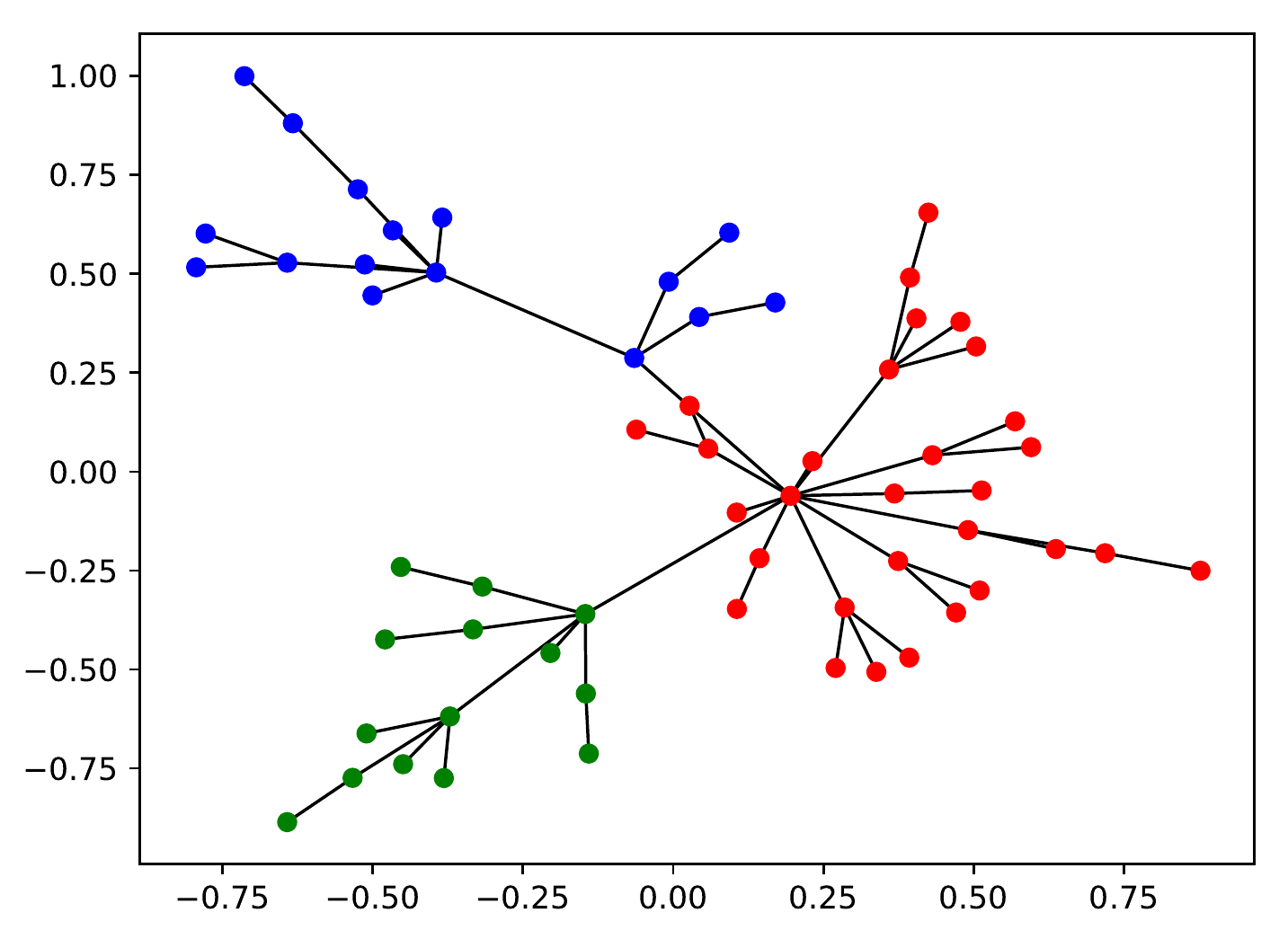}
\end{minipage}
}
\subfigure[BA60-20806.gexf]{
\begin{minipage}[t]{0.4\linewidth}
\centering
\includegraphics[width=1\linewidth]{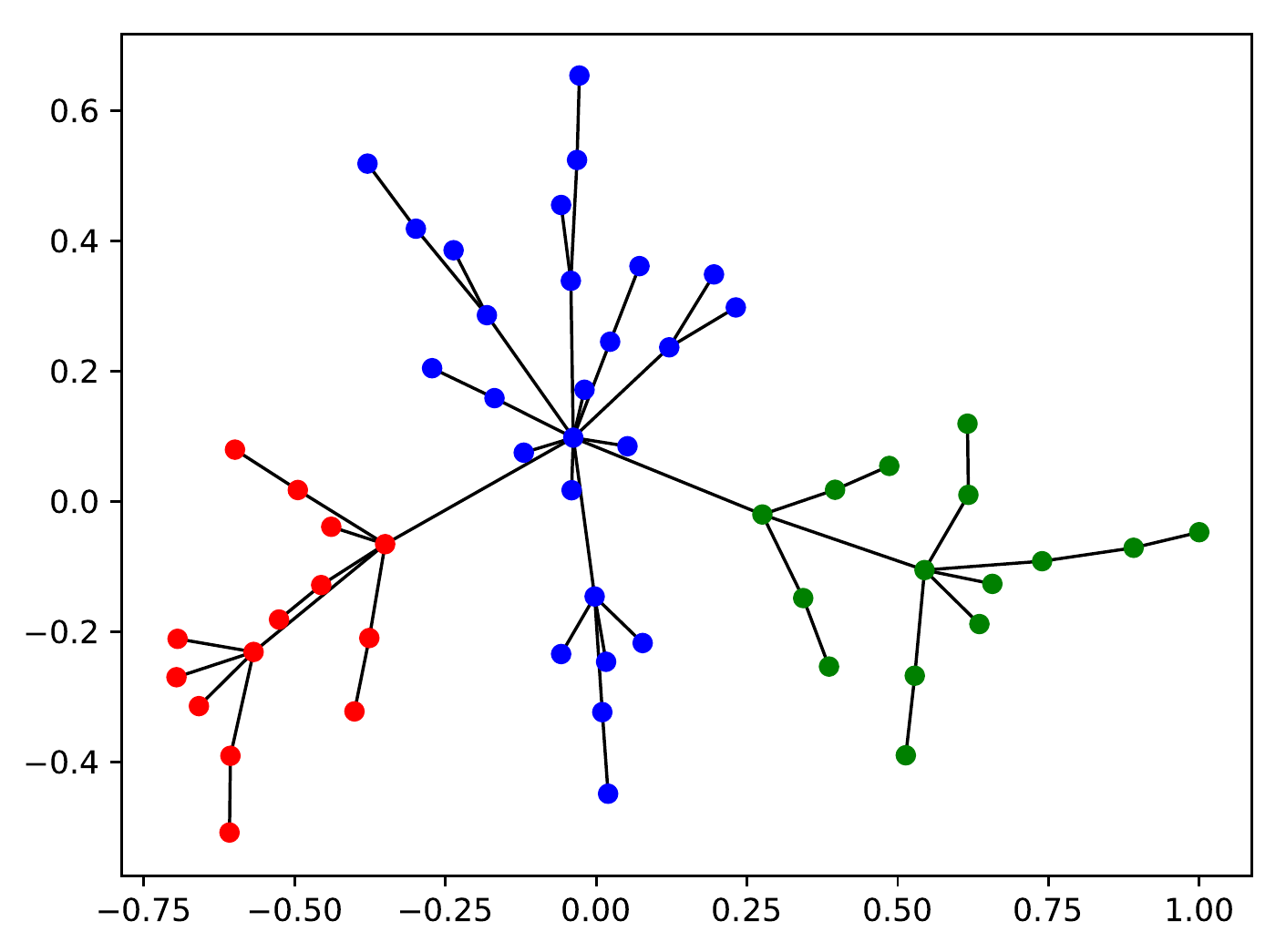}
\end{minipage}
}
\\
\subfigure[IMDB-X-370.gexf]{
\begin{minipage}[t]{0.4\linewidth}
\centering
\includegraphics[width=1\linewidth]{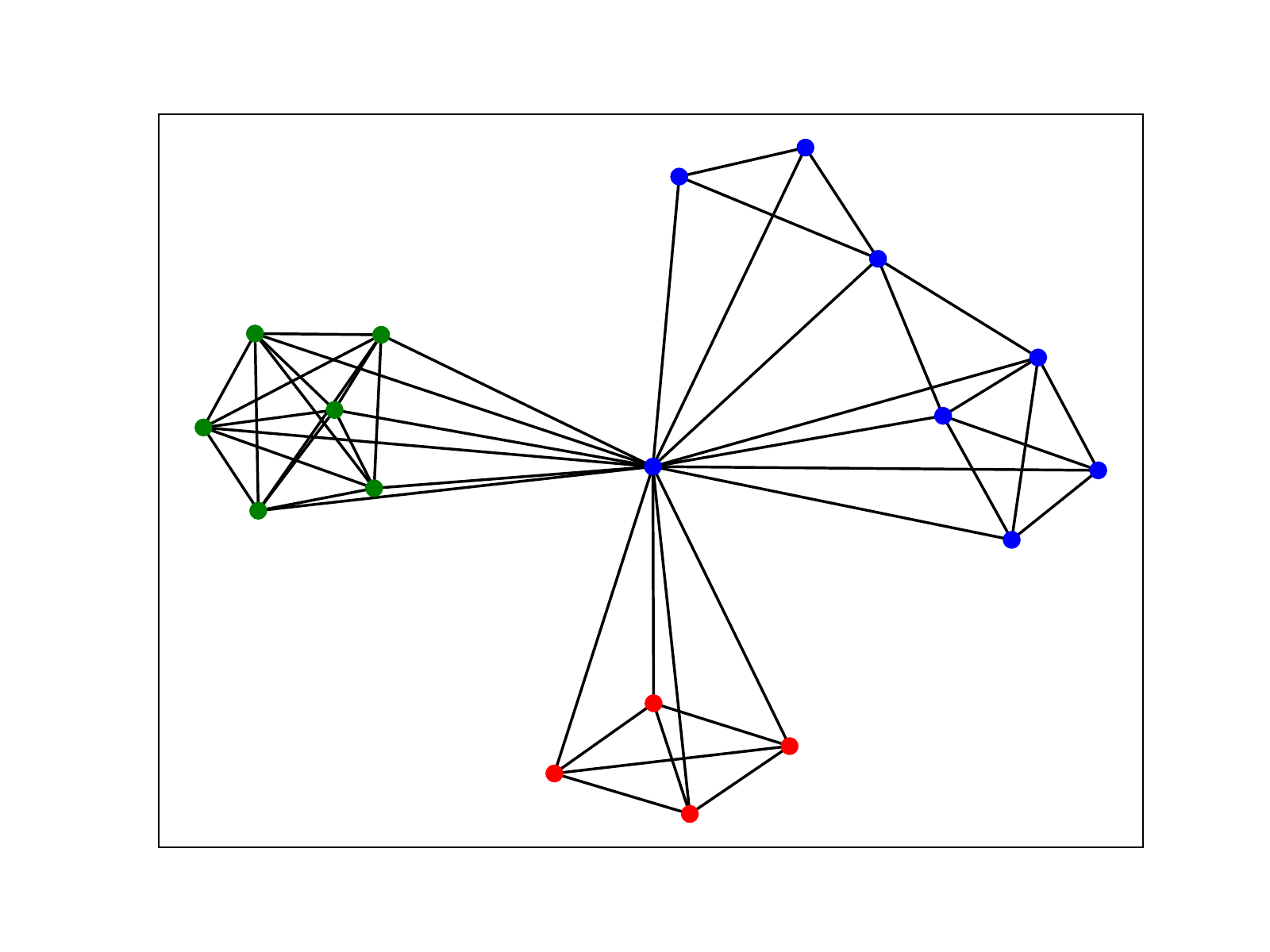}
\end{minipage}
}
\subfigure[IMDB-X-1456.gexf]{
\begin{minipage}[t]{0.4\linewidth}
\centering
\includegraphics[width=1\linewidth]{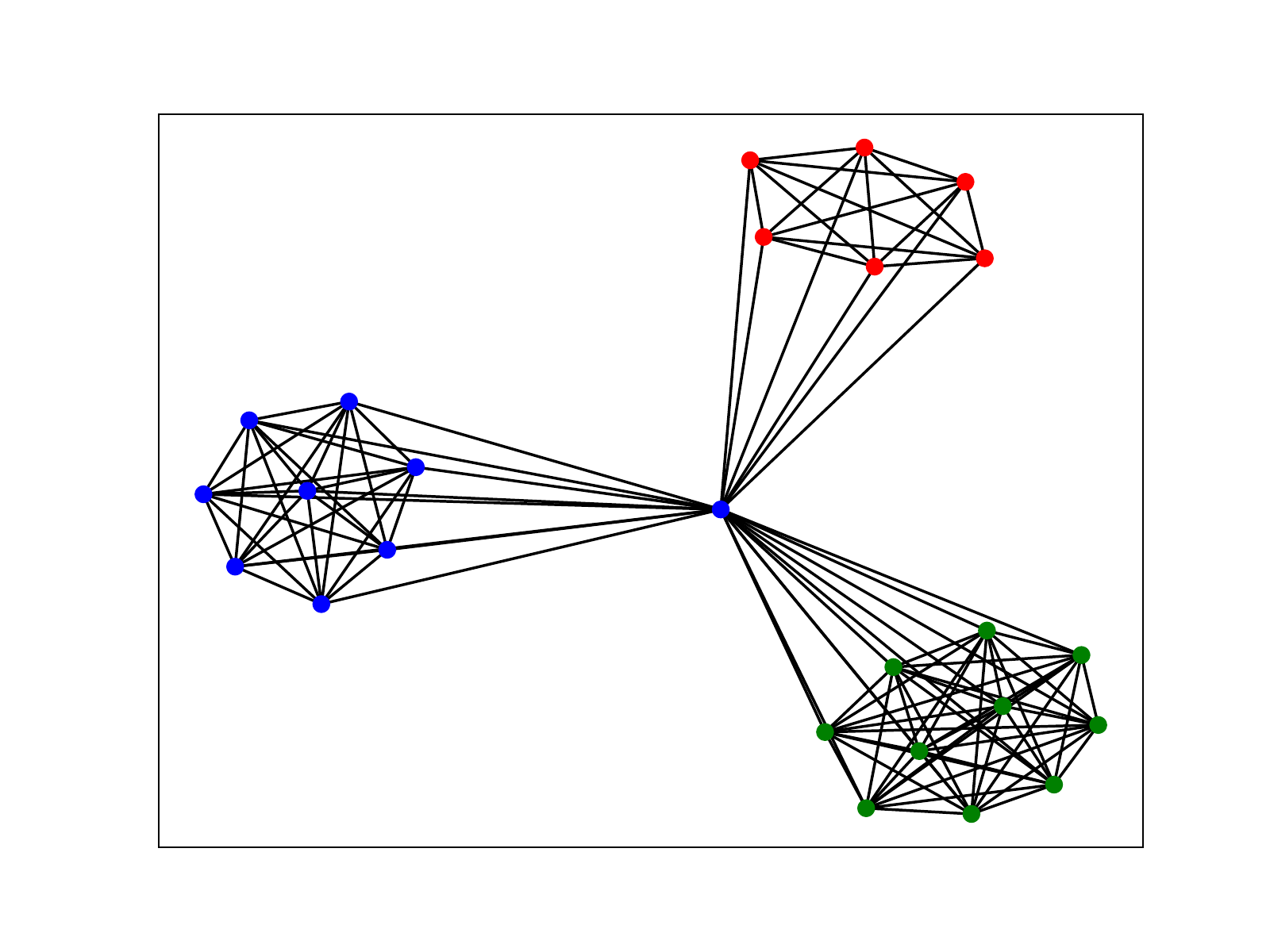}
\end{minipage}
}
\caption{Examples of graph partition from BA-60 and IMDB-X datasets. Different colors represent different subgraphs.}
\label{fig:Partition Case}
\vspace{-5mm}
\end{figure}

In view of this, we artificially made three BA-datasets, which have a large number of nodes and have graphs with obvious local structures by using the BA-model characteristics.
Table \ref{dataset} shows the comparison of different datasets for graph similarity computation.
Figure \ref{fig:Degree} shows the nodes degree distribution of BA-model datasets and IMDB-X.
From these charts, we can see that the average number of nodes to the number of edges in the BA-model datasets is approximately equal to 1.
Graphs are relatively sparse and suitable for extracting local structural features by graph partitioning.
The degree distribution indicates that most nodes have relatively low degrees, and only a few have high degrees.
These nodes have a greater probability of becoming the center node of the subgraph.
The partitioning results of the two graphs in the BA-60 dataset and two graphs in IMDB-X are shown in Figure \ref{fig:Partition Case}. Through graph partitioning, obvious local structural features can be extracted, which is also a characteristic of our BA-model datasets and IMDB-X.

\begin{figure}[t]
\centering
\includegraphics[width=0.75\linewidth]{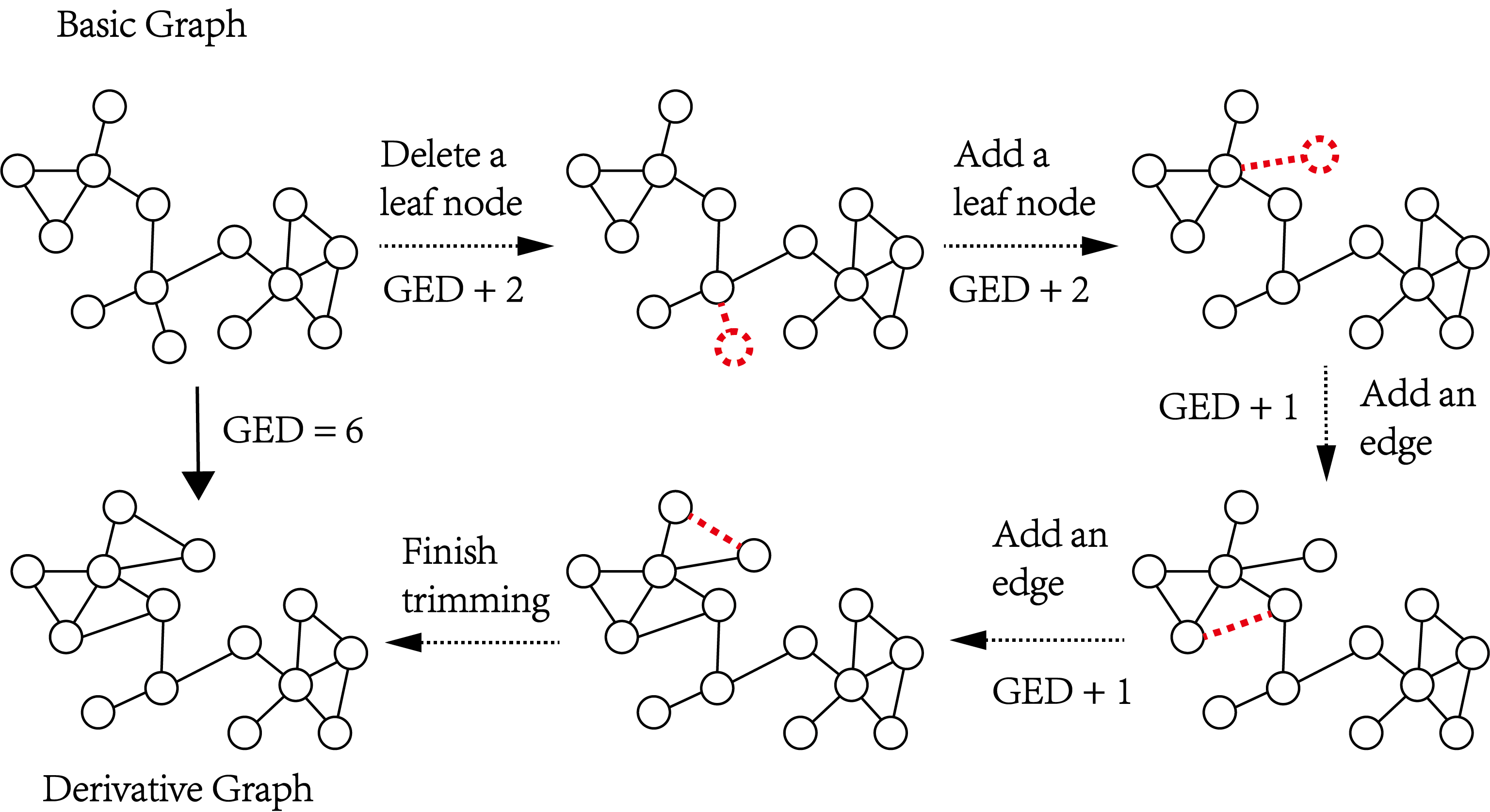}
\caption{Generate a derivative graph with a GED of 6 from the basic graph. It is not necessary to use all three methods in real trimming. Here is only one case.}
\label{fig:GED}
\vspace{-4mm}
\end{figure}
\subsection{Data Preprocessing and Ground-truth Generation}
For each dataset, we randomly split 60\%, 20\%, and 20\% of all graphs as the training set, validation set, and test set, respectively.
	
Due to the large number of nodes in our data set, \textit{A*} \cite{riesen2013novel} algorithm cannot be used to calculate the GED. We used the smallest distance calculated by three well-known approximation algorithms, \textit{Hungarian} \cite{kuhn1955hungarian,riesen2009approximate} and \textit{VJ} \cite{fankhauser2011speeding,jonker1987shortest}, and \textit{Beam} \cite{neuhaus2006fast}. However, these algorithms are also difficult to ensure a certain accuracy in this case. So we also added the GED value generated when trimming the graph as another evaluation indicator. When each graph is trimmed, we will get a GED value to record the number of
trimming steps.
Every time a leaf node is deleted in this experiment, the edge it connects will also be deleted. In this case, the GED between the derived graph and the basic graph increases by 2; and every time an edge is added, GED increases by 1.

As shown in Figure \ref{fig:GED}, when generating a derivative graph with a specific GED from the basic graph, we randomly select among the above three methods (randomly delete a leaf node, add a leaf node or add an edge) to generate a set of operations, and the sum of GED accumulated by all operations is the specific GED value.
We take the minimum value of the trimming GED and the calculated GED with these three algorithms as the final GED value. Here, the minimum value is taken instead of the average value because GED is the upper bound. The real GED value must be less than or equal to the GED PSimGNN-up only uses the subgraph-level embeddings and achieves the same level of evaluation results as other matching models, which proves the effectiveness of introducing subgraphs to help the large graphs similarity computation.
PSimGNN-$k$, which uses $k$ subgraph pairs for node-level comparison, achieves better results than PSimGNN-up on all evaluation metrics.
Our model, PSimGNN, consistently achieves the best or second-best under most evaluation metrics across the three datasets within the neural network-based methods.
In some ranking indicators ($\rho$,$\tau$, and $p@10$) of BA-100 and BA-200, although PSimGNN is not optimal, which may be caused by the randomness of graph partitioning, it still gets close to GSimCNN and GMN in performance.
This implies that our model introduces a more flexible framework and performs the same level of accuracy as other neural network-based models.
And as the number of subgraph pairs using node-level comparison increases, the model contains more information. The corresponding evaluation results become better, which is also in line with our expectations.
As mentioned before, for graphs in IMDB-X, there is no trimming steps.
The Beam algorithm's accuracy is higher than the other two, so most of the ground truth is the approximate GEDs calculated by Beam.
In this case, Beam shows the best performance in the table \ref{IMDB-X}, which also meets our expectations.
At the same time, \textit{PSimGNN} also shows the best performance compared with other deep learning methods on the IMDB dataset, which proves that the idea of graph partitioning is effective in the task of graph similarity computation.

As shown in Table \ref{time}, we recorded the running time for different models on test datasets.
It is worth noting here that these models are implemented in different ways, and the framework itself may cause the time issue.
In this case, it isn't easy to judge the computation cost directly by the running time is consumed.
For example, GCN-Mean and GCN-Max, although their computation cost is $O(N)$, to get the best performance, there are several layers of node embedding, so the time consumed is not the lowest. And GSimCNN, with $O(N^2)$ computation cost, has the fastest running time due to the optimization of CNN in the framework used.

However, since we used the GMN code for reference when implementing node-level interaction in PSimGNN, the two still have a comparative value.
By observing the running time of these two methods on three datasets of different scales, we can find that when the number of graph nodes is small, PSimGNN, which partitions first and then calculates the similarity, takes a long time due to a large number of steps. When the number of nodes is large, GMN directly interacts with the large graph at the node level with quadratic computation cost, which takes more time. Simultaneously, PSimGNN performs node interaction with the smaller graph, thus shortening the time, which proves our previous analysis of computation cost.
At the same time, the experimental results of PSimGNN-up, PSimGNN-k, and PSimGNN also indicate that the more subgraphs involved in the computation, the longer it takes.
d here.
There are no trimming steps for graphs in IMDB-X, so we only these three well-known approximation algorithms to get the ground truth GED.

In order to convert the calculated GED into the similarity score required by our model, we first normalize the GED by $nGED(G_1,G_2)=\frac{GED(G_1,G_2)}{(|G_1|+|G_2|)/2}$, where $|G_i|$ represents the total number of nodes in graph $G_i$. Then use the exponential function $f(x)=e^{-x}$ to map the normalized GED to between 0 and 1 to represent the pair's graph similarity. Here we can see that the more similar the graph, the smaller the GED, and the more similarity tends to 1.

\subsection{Baseline Methods}
Our baseline includes three categories of methods, fast approximate GED calculation algorithms, graph embedding based models, and graph matching network-based models.

\begin{itemize}
\item The first category of baseline includes three classic algorithms for GED calculation. (1) \textit{Hungarian} \cite{kuhn1955hungarian,riesen2009approximate} is a cubic-time algorithms based on the Hungarian Algorithm for bipartite graph matching. (2) \textit{VJ} \cite{fankhauser2011speeding,jonker1987shortest} is also a cubic-time algorithms based on the algorithm of Volgenant and Jonker. (3) \textit{Beam search} (\textit{Beam}) \cite{neuhaus2006fast}. The equivalent variable of the \textit{A*} algorithm is sub-exponential time.
\item The second category of baseline includes two graph embedding based models, GCN-Mean and GCN-Max \cite{defferrard2016convolutional}. They all embed graphs into vectors using GCN and then use the similarities calculated by these vectors as the similarities of these graph pairs.
\item The third category of baseline includes three graph matching network-based models. (1) SimGNN \cite{bai2019simgnn} and (2) GSimCNN \cite{bai2018convolutional} combine the embedding of the whole graph and node-level comparison. (3) GMN \cite{li2019graph} uses the comparison node information within and between graphs to calculate similarity.
\end{itemize}

Our method also belongs to the third category of methods, using graph matching based networks to calculate the similarities of graph pairs.

\subsection{Parameter Settings}
For the architecture of our model, PSimGNN, we partition each large graph into $k$ (here $k$=3) subgraphs.
Among the nine subgraph pairs, 0, 3, and 9 subgraph pairs with the highest similarity scores are selected for node-level comparison, respectively.
Here we call them PSimGNN-up (only subgraph-level interactions are involved in the computation), PSimGNN-$k$ ($k$ or three pairs of subgraphs participate in the node-level comparison), and PSimGNN (all or nine pairs of subgraphs participate in the node-level comparison).
It is worth mentioning that we do not perform graph partition for graphs with very few nodes, such as the graphs in the AIDS and LINUX datasets.

We set the number of GIN \cite{xu2018powerful} layer to 3, and use Parametric Rectified Linear Unit (PReLU) \cite{he2015delving} as the activation function. For the initial node representations, we adopt the constant encoding scheme for BA-datasets since their nodes are unlabeled, as mentioned in Section 3.2.1. The dimensions of the 1st, 2nd, and 3rd layer of GIN's output are 64, 32, and 16, respectively.
We use a fully connected layer to reduce the similarity vectors' dimension obtained at the subgraph-level interaction from 9 to 8, and another fully connected layer to change the dimension of the similarity vector after the node-level comparison from 3 to 8. Finally, four fully connected layers are used to reduce the dimension of the concatenated results from the subgraph-level interaction and the node-level comparison module, from 16 to 8, 8 to 4, 4 to 2, and 2 to 1.

For training, we set the batch size to 128, use the Adam algorithm \cite{kingma2014adam} for optimization, and set the initial learning rate to 0.001. We set the number of training iterations to 2000 and choose the best model based on the lowest validation loss.

\subsection{Evaluation Metrics}

We used two metrics to evaluate the similarity computation results of this model. \textit{Mean Squared Error (MSE)}. MSE measures the average squared difference between all the calculated similarities and the ground-truth similarities. \textit{Mean Absolute Error (MAE)}. MAE measure the averaged value of the absolute deviation of all the calculated similarities from the ground-truth similarities.

For the ranking results, we also use \textit{Spearman's Rank Correlation Coefficient ($\rho$)} \cite{spearman1961proof} and \textit{Kendall's Rank Correlation Coefficient ($\tau$)} \cite{kendall1938new} to evaluate how well the predicted ranking results match the true ranking results.
\textit{Precision at k} $(p@k)$ is computed by taking the intersection of the predicted top $k$ results and the ground truth top $k$ results divided by $k$. Compared with $p@k$, $\rho$ and $\tau$ can better reflect the global ranking results instead of focusing on the top $k$ results.

\begin{table}[t]
\renewcommand\arraystretch{1.1}
\caption{Results on three common datasets ($\bm{10^{-2}}$). The best results are bolded.}
\label{common}
\centering
\scalebox{0.75}{
\begin{tabular}{cccccccccc}
\hline
\multirow{2}{*}{Method} &               & AIDS           &                &               & lINUX          &                &               & IMDB           &                \\
                        & mse           & $\tau$              & p@10           & mse           & $\tau$              & p@10           & mse           & $\tau$ & p@10           \\ \hline
hungarian               & 2.53          & 37.80          & 36.00          & 2.98          & 51.70          & 91.30          & 0.18          & 87.20          & 82.50          \\
vj                      & 2.92          & 38.30          & 31.00          & 6.39          & 45.00          & 28.70          & 0.18          & 87.40          & 81.50          \\
beam                    & 1.21          & 46.30          & 48.10          & 0.93          & 71.40          & 97.30          & 0.24          & 83.70          & 80.30          \\ \hline
GCN-Mean                & 0.34          & 50.10          & 18.60          & 0.85          & 42.40          & 14.10          & 0.69          & 30.70          & 20.00          \\
GCN-Max                 & 0.37          & 48.00          & 19.50          & 0.64          & 49.50          & 43.70          & 0.51          & 34.20          & 42.50          \\ \hline
SimGNN                  & 0.12          & 69.00          & 42.10          & 0.15          & 83.00          & 94.20          & 0.13          & 77.00          & 75.90          \\
GSimCNN                 & 0.08          & 72.40          & 52.10          & 0.10          & \textbf{96.20} & \textbf{99.20} & 0.08          & \textbf{84.70} & 82.80          \\
GMN                     & \textbf{0.07} & \textbf{73.20} & 52.30          & \textbf{0.08} & 95.70          & 96.80          & 0.08          & 81.80          & 82.30          \\
PSimGNN                 & 0.11          & 70.10          & \textbf{53.40} & 0.12          & 91.50          & 97.70          & \textbf{0.07} & 82.20          & \textbf{83.10} \\ \hline
\end{tabular}}
\vspace{-4mm}
\end{table}

\begin{table}[t]
\renewcommand\arraystretch{1.1}
\caption{Results on BA-60 dataset ($\bm{10^{-2}}$). The best results of the neural network-based models, as well as the traditional methods that exceed these results are bolded.}
\label{BA-60}
\centering
\scalebox{0.75}{
\begin{tabular}{|c|cccccc|}
\hline
Method &MSE &MAE &$\rho$ &$\tau$ &\textbf{p@10} &\textbf{p@20}
\\ \hline
hungarian &18.62 &33.22 &75.98 &57.72 &74.25 &84.75 \\
vj  &25.87 &39.48 &3.29 &2.29 &35.00 &50.50 \\
beam  &5.88 &12.93 &\textbf{85.80} &\textbf{74.34} &67.75 &90.00
\\\hline
GCN-Mean &0.58 &5.39 &75.64 &53.29 &58.00 &86.88 \\
GCN-Max  &1.37 &9.14 &74.61 &52.30 &54.50 &86.62
\\ \hline
SimGNN &0.78 &6.58 &77.30 &56.78 &71.00 &88.87 \\
GSimCNN &0.60 &5.61 &80.78 &60.47 &67.75 &90.50 \\
GMN &0.27 &3.82 &76.36 &54.67 &60.00 &89.00 \\
PSimGNN-up &0.44 &4.80 &78.92 &57.63 &59.50 &88.37 \\
PSimGNN-$k$ &0.32 &4.07 &80.43 &60.31 &70.50 &88.00 \\
PSimGNN &\textbf{0.20} &\textbf{3.39} &\textbf{84.49} &\textbf{66.15} &\textbf{78.50} &\textbf{91.87}
\\ \hline
\end{tabular}}
\vspace{-4mm}
\end{table}

\begin{table}[t]
\renewcommand\arraystretch{1.1}
\caption{Results on BA-100 dataset ($\bm{10^{-2}}$).}
\label{BA-100}
\centering
\scalebox{0.75}{
\begin{tabular}{|c|cccccc|}
\hline
Method &MSE &MAE &$\rho$ &$\tau$ &\textbf{p@10} &\textbf{p@20}
\\ \hline
hungarian &20.54 &34.38 &81.10 &60.36 &61.00 &99.00 \\
vj &27.39 &40.46 &58.37 &41.56 &46.25 &82.62 \\
beam &11.40 &20.68 &78.67 &\textbf{62.83} &62.75 &90.00
\\\hline
GCN-Mean &1.25 &9.09 &76.39 &53.38 &56.50 &\textbf{100} \\
GCN-Max  &1.20 &8.54 &76.17 &53.04 &52.50 &99.88
\\ \hline
SimGNN &0.80 &6.93 &76.37 &53.83 &58.00 &\textbf{100.00}\\
GSimCNN &0.23 &3.25 &\textbf{82.33} &\textbf{61.69} &\textbf{67.00} &\textbf{100.00}\\
GMN &0.15 &2.71 &77.22 &54.50 &53.25 &\textbf{100.00}\\
PSimGNN-up &0.50 &4.24 &77.71 &55.33 &53.50 &\textbf{100.00}\\
PSimGNN-$k$ &0.12 &2.51 &79.65 &57.81 &57.75 &\textbf{100.00}\\
PSimGNN &\textbf{0.11} &\textbf{2.41} &80.14 &58.44 &61.25 &\textbf{100.00}
\\ \hline
\end{tabular}}
\vspace{-4mm}
\end{table}

\begin{table}[t]
\renewcommand\arraystretch{1.1}
\caption{Results on BA-200 dataset ($\bm{10^{-2}}$).}
\label{BA-200}
\centering
\scalebox{0.75}{
\begin{tabular}{|c|cccccc|}
\hline
Method &MSE &MAE &$\rho$ &$\tau$ &\textbf{p@10} &\textbf{p@20}
\\ \hline
hungarian &25.91 &37.94 &79.38 &58.10 &\textbf{64.25} &94.00 \\
vj &31.44 &42.68 &61.91 &43.10 &48.50 &80.38 \\
beam &18.60 &28.79 &77.24 &\textbf{65.21} &56.00 &83.50
\\\hline
GCN-Mean &2.37 &12.78 &73.47 &49.46 &50.00 &95.00 \\
GCN-Max &2.28 &10.76 &74.99 &51.69 &53.75 &94.25
\\ \hline
SimGNN &0.84 &6.19 &73.47 &48.89 &52.75 &95.13 \\
GSimCNN &0.32 &3.58 &\textbf{79.68} &56.82 &59.00 &95.00\\
GMN &0.12 &2.66 &79.58 &\textbf{57.87} &\textbf{60.25} &95.00\\
PSimGNN-up &0.08 &4.53 &74.95 &51.58 &46.75 &95.13 \\
PSimGNN-$k$ &0.07 &2.14 &76.36 &53.29 &52.50 &96.00 \\
PSimGNN &\textbf{0.06} &\textbf{1.96} &79.16 &57.24 &55.75 &\textbf{97.63}
\\ \hline
\end{tabular}}
\vspace{-4mm}
\end{table}

\begin{table}[t]
\renewcommand\arraystretch{1.1}
\caption{Results on IMDB-X dataset ($\bm{10^{-2}}$).}
\label{IMDB-X}
\centering
\scalebox{0.75}{
\begin{tabular}{|c|cccccc|}
\hline
Method &MSE &MAE &$\rho$ &$\tau$ &\textbf{p@10} &\textbf{p@20}
\\ \hline
hungarian & 0.27 & 1.66 & 93.09 & 83.18 & 74.09 & 80.91 \\
vj & 0.77 & 2.27 & 93.32 & 83.34 & 74.50 & 81.48 \\
beam & \textbf{0.04} & \textbf{0.49} & \textbf{96.01} & \textbf{90.40} & \textbf{90.91} & \textbf{90.68}
\\\hline
GCN-Mean & 2.22 & 5.54 & 46.64 & 36.20 & 48.64 & 70.22 \\
GCN-Max & 4.71 & 12.32 & 24.62 & 17.36 & 39.09 & 44.55
\\ \hline
SimGNN & 0.74 & 3.37 & 52.70 & 39.35 & 55.68 & 61.47 \\
GSimCNN & 0.50 & 3.04 & 66.26 & 49.87 & 62.05 & 64.09 \\
GMN & 0.38 & 2.73 & 69.59 & 55.38 & 65.68 & 71.82 \\
PSimGNN-up & 0.82 & 4.41 & 53.74 & 42.38 & 59.00 & 61.70 \\
PSimGNN-$k$ & 0.42 & 3.01 & 68.23 & 51.42 & 62.50 & 64.43 \\
PSimGNN & \textbf{0.31} & \textbf{2.51} & \textbf{72.34} & \textbf{60.31} & \textbf{68.18} & \textbf{73.86}
\\ \hline
\end{tabular}}
\vspace{-4mm}
\end{table}

\begin{table}[t]
\renewcommand\arraystretch{1.1}
\caption{Results for average time consumption on one pair of graphs in milliseconds.}
\label{time}
\centering
\scalebox{0.75}{
\begin{tabular}{|c|ccc|}
\hline
Method &BA-60 &BA-100 &BA-200
\\ \hline
hungarian &229 &331 &1192 \\
vj &221 &308 &1010 \\
beam &177 &254 &737
\\\hline
GCN-Mean &6.4 &8.7 &10.2\\
GCN-Max &7.1 &9.2 &11.3
\\ \hline
SimGNN &4.4 &5.6 &6.9\\
GSimCNN &2.5 &3.1 &5.6\\
GMN &9.4 &13.8 &37.5\\
PSimGNN-up &3.8 &5.0 &9.4\\
PSimGNN-$k$ &8.1 &10.6 &15.6\\
PSimGNN &15.6 &20.0 &30.0
\\ \hline
\end{tabular}}
\vspace{-4mm}
\end{table}

\subsection{Results and Analysis}

The experimental results on these datasets can be found in Table \ref{common}, \ref{BA-60}, \ref{BA-100}, \ref{BA-200} and \ref{IMDB-X}.
Table \ref{common} shows that on small graphs where the local features are not obvious enough, PSimGNN can show comparable performance to other matching models.
This proves that the framework is suitable for small graphs, with great scalability.
The ranking results of \textit{VJ} on the BA-60 dataset is extremely poor, and these three traditional methods also have very high \emph{MSE} and \emph{MAE}. These results show the limitations of traditional methods for graphs with a large number of nodes.
However, as far as the index $\tau$ is concerned, the optimal value of \textit{beam} continues to exceed the neural network methods.
This may be due to the way that \textit{beam} directly acts on edges and nodes in the BA graph can better distinguish the distance between query graphs and graphs in the database, thus having advantages in the ranking.
As for the BA-100 dataset, $p@20$ is 100\%. This is because when randomly dividing the test dataset of 40 graphs, there are exactly 20 graphs from the basic graph 1, and the other 20 graphs are from the basic graph 2. Being able to distinguish these graphs correctly also proves the excellent performance of the neural network-based models.

For all datasets, the GCN-Mean and GCN-Max results are worse than any matching model in terms of most evaluation indicators.
When the number of nodes per graph increases, the limitation of using one vector to characterize the entire graph is more obvious, and the results are worse, which also confirms our previous analysis in \hyperref[subsec: Graph Partitioning]{Section 3.2}.

PSimGNN-up only uses the subgraph-level embeddings and achieves the same level of evaluation results as other matching models, which proves the effectiveness of introducing subgraphs to help the large graphs similarity computation.
PSimGNN-$k$, which uses $k$ subgraph pairs for node-level comparison, achieves better results than PSimGNN-up on all evaluation metrics.
Our model, PSimGNN, consistently achieves the best or second-best under most evaluation metrics across the three datasets within the neural network-based methods.
In some ranking indicators ($\rho$,$\tau$, and $p@10$) of BA-100 and BA-200, although PSimGNN is not optimal, which may be caused by the randomness of graph partitioning, it still gets close to GSimCNN and GMN in performance.
This implies that our model introduces a more flexible framework and performs the same level of accuracy as other neural network-based models.
And as the number of subgraph pairs using node-level comparison increases, the model contains more information. The corresponding evaluation results become better, which is also in line with our expectations.
As mentioned before, for graphs in IMDB-X, there is no trimming steps.
The Beam algorithm's accuracy is higher than the other two, so most of the ground truth is the approximate GEDs calculated by Beam.
In this case, Beam shows the best performance in the table \ref{IMDB-X}, which also meets our expectations.
At the same time, \textit{PSimGNN} also shows the best performance compared with other deep learning methods on the IMDB dataset, which proves that the idea of graph partitioning is effective in the task of graph similarity computation.

As shown in Table \ref{time}, we recorded the running time for different models on test datasets.
It is worth noting here that these models are implemented in different ways, and the framework itself may cause the time issue.
In this case, it isn't easy to judge the computation cost directly by the running time is consumed.
For example, GCN-Mean and GCN-Max, although their computation cost is $O(N)$, to get the best performance, there are several layers of node embedding, so the time consumed is not the lowest. And GSimCNN, with $O(N^2)$ computation cost, has the fastest running time due to the optimization of CNN in the framework used.

However, since we used the GMN code for reference when implementing node-level interaction in PSimGNN, the two still have a comparative value.
By observing the running time of these two methods on three datasets of different scales, we can find that when the number of graph nodes is small, PSimGNN, which partitions first and then calculates the similarity, takes a long time due to a large number of steps. When the number of nodes is large, GMN directly interacts with the large graph at the node level with quadratic computation cost, which takes more time. Simultaneously, PSimGNN performs node interaction with the smaller graph, thus shortening the time, which proves our previous analysis of computation cost.
At the same time, the experimental results of PSimGNN-up, PSimGNN-k, and PSimGNN also indicate that the more subgraphs involved in the computation, the longer it takes.

\begin{table}[t]
\renewcommand\arraystretch{1.1}
\caption{Ablation Study on BA-60 dataset ($\bm{10^{-2}}$). The best results are bolded.}
\centering
\scalebox{0.75}{
\begin{tabular}{|c|cccccc|}
\hline
Method &MSE &MAE &$\rho$ &$\tau$ &\textbf{p@10} &\textbf{p@20}
\\ \hline
PSimGNN\_sub\_att &0.31 &3.91 &83.17 &62.50 &72.75 &90.37 \\
PSimGNN\_cross\_att &0.29 &3.88 &82.29 &61.30 &73.50 &90.87 \\
PSimGNN\_cross &0.61 &5.72 &79.83 &57.82 &71.50 &89.37 \\
PSimGNN\_within &0.57 &5.57 &80.92 &58.36 &69.50 &90.00 \\
PSimGNN &\textbf{0.20} &\textbf{3.39} &\textbf{84.49} &\textbf{66.15} &\textbf{78.50} &\textbf{91.87}
\\ \hline
\end{tabular}}
\vspace{-5mm}
\label{ablation}
\end{table}

\subsection{Ablation Study}

We evaluated how each of the components of PSimGNN affects the results. 
We report the results on the BA-60 dataset after removing a specific part. 
PSimGNN\_sub\_att represents our model without the attention mechanism in subgraph embedding.
Here we use the average pooling method as an alternative.
PSimGNN\_cross\_att indicates our model without the attention mechanism in node-level comparison.
We also give each node the same weight.
PSimGNN\_cross and PSimGNN\_within, respectively, means that our model removes the interaction within the subgraph and between subgraphs.
PSimGNN is the complete model we proposed.
It can be seen from Table \ref{ablation} that regardless of removing each component, the performance of PSimGNN shows attenuation, and removing the attention mechanism has less impact on performance than the other two. 
This makes intuitive sense. 
Because for the latter two cases, they will lose a lot of information within or between graphs, which is not conducive to the high-quality embedding.
This also further proves that the aggregation methods and attention mechanism are effective. 
With their joint contribution, PSimGNN can perform well in the graph similarity computation task.

\begin{figure}[t]
\centering
\subfigure[Partition Number $k$]{
\begin{minipage}[t]{0.4\linewidth}
\centering
\includegraphics[width=1\linewidth]{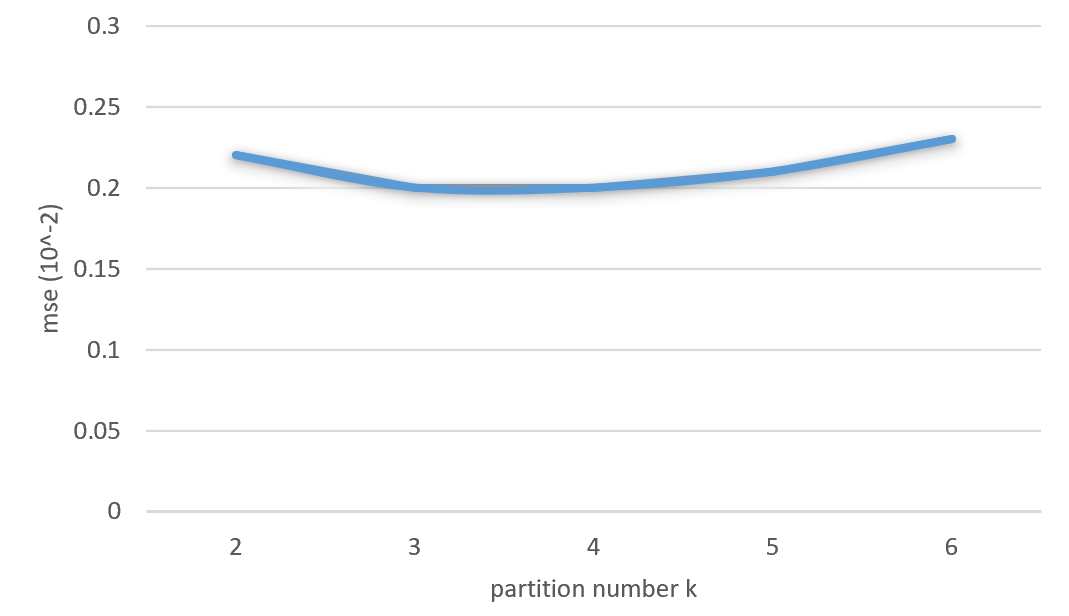}
\end{minipage}
\label{sensitivity-k}
}
\subfigure[Embedding Dimension $D$]{
\begin{minipage}[t]{0.4\linewidth}
\centering
\includegraphics[width=1\linewidth]{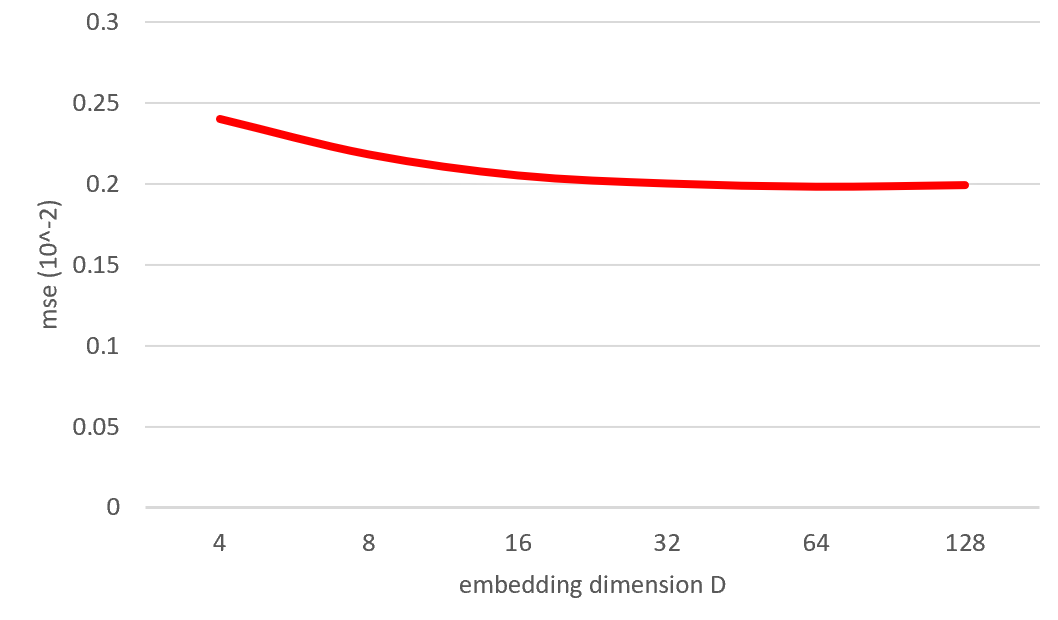}
\end{minipage}
\label{sensitivity-D}
}
\caption{Mean squared error with respect to the partition number $k$ and the dimensions $D$ of subgraph-level embeddings.}
\label{sensitivity}
\vspace{-5mm}
\end{figure}

\subsection{Parameter Sensitivity}

In this part, we analyze the effect of partition number $k$ and subgraph embedding dimension $D$ on the performance of PSimGNN. 
We report the mean squared error on the BA-60 dataset.
The lower the mse, the better the performance.
Figure \ref{sensitivity-k} shows the performance comparison of the partition number $k$.
As the number of partition subgraphs increases, the performance will first rise and then decline. 
This is because too few subgraphs cannot make full use of the graph's local features, and too many subgraphs will destroy the overall structure of the graph. Therefore, we need to choose a reasonable number of subgraphs to achieve optimal performance.
Figure \ref{sensitivity-D} presents the performance comparison of the subgraph embedding dimension. 
In general, with the increase of the embedding dimension, the performance will increase. 
It makes intuitive sense since larger subgraph embedding dimension $D$ can provide PSimGNN more capacity to represent subgraphs.
However, when $D$ reaches a certain level, around 32 here, performance growth becomes slow.
Therefore, we need to find a proper length of embedding to balance the trade-off between the performance and the complexity. 

\section{Conclusion and Future Directions}
\label{cha:conclusion}
We are at the intersection of graph neural network, graph similarity computation, and graph partition. We are taking the first step towards large graph similarity computation via graph partition and a novel neural network-based approach PSimGNN. The proposed method's central idea is to solve the problem of large graph similarity computation from the perspective of subgraphs, which takes any two graphs as input and outputs their similarity score. The experimental results show that PSimGNN achieves competitive accuracy and computation cost by introducing graph partitioning.

There are several directions to go for future work: 
(1) State-of-the-art graph partitioning methods usually consist of three main phases: coarsening, initial partitioning, and uncoarsening. It's very interesting to explore whether we can only deploy the coarsening stage on large graphs and split each large graph into some soft clusters (partitioning results is hard clusters). Then similarity computation based on these soft clusters may further reduce computational computation cost involved in the node-node similarity computation.
(2) Introducing a mechanism to deal with edge attributes \cite{zhang2015comprehensive} is promising in some applications.
In chemistry, atomic properties and bonds of a chemical compound are usually labeled, so it is useful to incorporate edge labels into our model.
(3) Given the constraint that the exact GEDs for large graphs cannot be computed, we can only use approximate GEDs. When the number of graph nodes is further larger, the approximation algorithms become even less accurate. It would be interesting to see how the learned model generalizes to larger graphs trained only on the exact GEDs between partitioned subgraphs or other small graph datasets.

\section{Acknowledgments}
We thank Yunsheng Bai and Derek Xu for valuable discussions. This work is supported in part by the National Key Research and Development Program (No.2019YFB2102600), the Fundamental Research Funds for the Central Universities (No.2020CDJQY-A005,No.2019CDQYRJ006), the National Natural Science Foundation of China (No.62002035,No.61702062), the Natural Science Foundation of Chongqing, China (No.cstc2020jcyj-bshX0034) and the Science and Technology Major Project of Ningbo City (No. 2018B10047).

\bibliographystyle{cas-model2-names}

\bibliography{cas-sc-template}


\bio{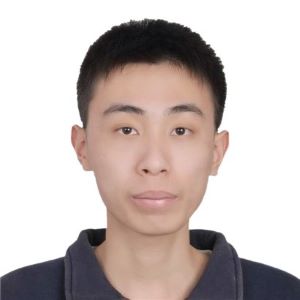}
\textbf{Haoyan Xu} received his B.S. in July, 2020 at Zhejiang University, College of Control Science and Engineering. His research interests lie in the area of Graph Representation Learning, Time Series Analysis, Robot Learning and Microfluidics. He is particular interested in graph neural networks, with their applications in language processing, graph mining, etc. What's more, he likes studying micro-scale robotics and bioinspired robotics and exploring fundamental questions related with them.
\vspace{1\baselineskip}
\endbio

\bio{author/Ziheng.png}
\textbf{Ziheng Duan} received his B.S. in July, 2020 at Zhejiang University, College of Control Science and Engineering. His research interests lie in the area of Machine Learning, Graph Representation Learning, Time Series Analysis, espeically the interaction of them.
\vspace{4\baselineskip}
\endbio

\bio{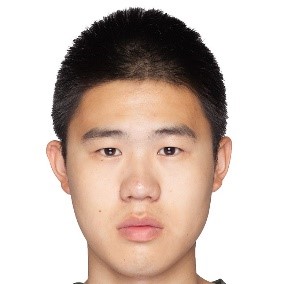}
\textbf{Jie Feng} is currently a senior student at Zhejiang University, College of Control Science and Engineering, who will receive his B.S. in June. 2021. His research interests include artificial intelligence and robotics.
\vspace{4\baselineskip}
\endbio

\bio{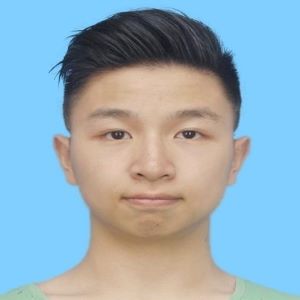}
\textbf{Runjian Chen} received the B.S. degree from the Department of Control Science and Engineering, Zhejiang University, Hangzhou, China, in 2020. His latest research interests include machine learning and its application in robotics and computer vision.
\vspace{4\baselineskip}
\endbio

\bio{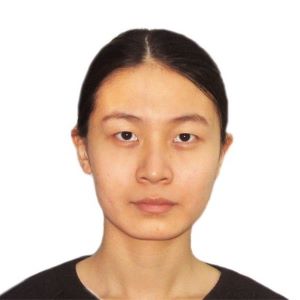}
\textbf{Qianru Zhang} is studying at Harbin Institute of Technology and will receive her bachelor degree in 2021. Her research interests lie in Machine Learning, Time Series Analysis and Graph Representation Learning.
\vspace{4\baselineskip}
\endbio

\bio{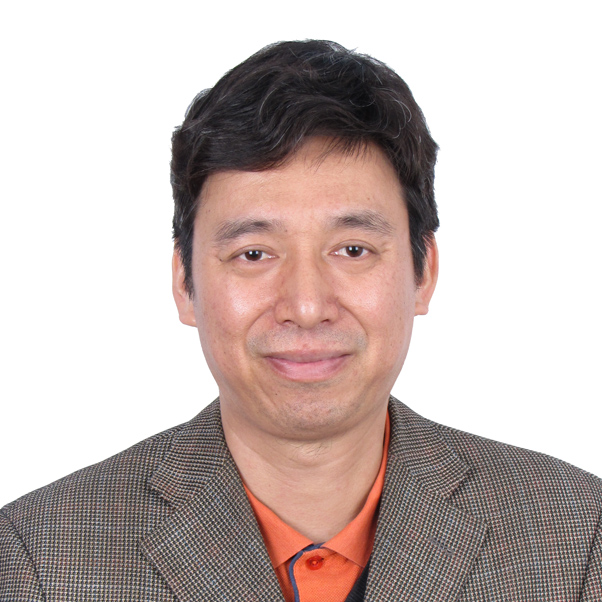}
\textbf{Zhongbin Xu} is a professor in Energy Engineering College at Zhejiang University. He received his PhD from South China University of Technology in 2001. He was a visiting scholar at the University of Cambridge (2008-2009) and Harvard University (2014-2015). His current research interests include intelligent microstructure manufacturing, polymer processing engineering and microfluidics. His group has done lots of work in the application of industrial intelligence.
\vspace{4\baselineskip}
\endbio

\bio{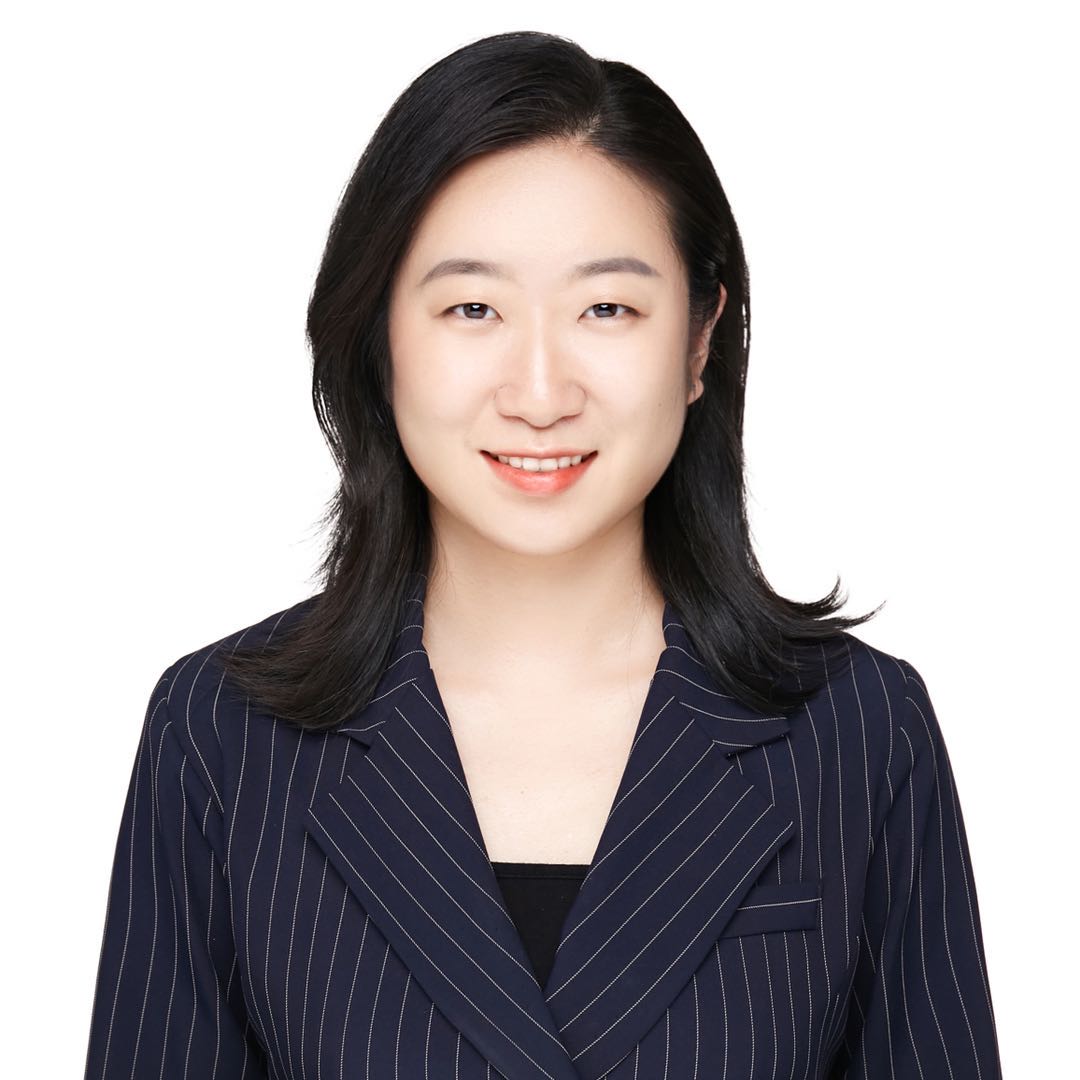}
\textbf{Yueyang Wang} received the B.E. degree from the Software Institute, Nanjing University, Nanjing, China, and the Ph.D. degree from Zhejiang University, Hangzhou, China. She is currently a Lecturer with the School of Big Data and Software Engineering, Chongqing University. Her research interests include social network analysis and data mining.
\vspace{4\baselineskip}
\endbio

\end{document}